\documentclass[sigconf]{acmart}

\usepackage{algorithm}
\usepackage{algpseudocode}  
\usepackage{amsmath}  
\usepackage{multirow, makecell}
\usepackage{booktabs}
\usepackage{graphicx}
\usepackage{subfigure}
\usepackage{pifont}
\usepackage{color}

\usepackage{amssymb}
\usepackage{amsthm}

\AtBeginDocument{%
  \providecommand\BibTeX{{%
    \normalfont B\kern-0.5em{\scshape i\kern-0.25em b}\kern-0.8em\TeX}}}

\copyrightyear{2022}
\acmYear{2022}
\setcopyright{acmcopyright}\acmConference[KDD '22]{Proceedings of the 28th ACM SIGKDD Conference on Knowledge Discovery and Data Mining}{August 14--18, 2022}{Washington, DC, USA}
\acmBooktitle{Proceedings of the 28th ACM SIGKDD Conference on Knowledge Discovery and Data Mining (KDD '22), August 14--18, 2022, Washington, DC, USA}
\acmPrice{15.00}
\acmDOI{10.1145/3534678.3539320}
\acmISBN{978-1-4503-9385-0/22/08}

\begin{document}

%%
%% The "title" command has an optional parameter,
%% allowing the author to define a "short title" to be used in page headers.
\title{FreeKD: Free-direction Knowledge Distillation for Graph Neural Networks}

%%
%% The "author" command and its associated commands are used to define
%% the authors and their affiliations.
%% Of note is the shared affiliation of the first two authors, and the
%% "authornote" and "authornotemark" commands
%% used to denote shared contribution to the research.

%\author{Lars Th{\o}rv{\"a}ld}
%\affiliation{%
  %\institution{The Th{\o}rv{\"a}ld Group}
  %\streetaddress{1 Th{\o}rv{\"a}ld Circle}
 % \city{Hekla}
 % \country{Iceland}}
%\email{larst@affiliation.org}

\author{Kaituo Feng}
\affiliation{%
 \institution{Beijing Institute of Technology}
 \city{Beijing}
 \country{China}
}
\email{kaituofeng@gmail.com}

\author{Changsheng Li}
\authornote{Corresponding author.}
\affiliation{%
 \institution{Beijing Institute of Technology}
 \city{Beijing}
 \country{China}
}
\email{lcs@bit.edu.cn}

\author{Ye Yuan}
\affiliation{%
 \institution{Beijing Institute of Technology}
 \city{Beijing}
 \country{China}
}
\email{yuan-ye@bit.edu.cn}

\author{Guoren Wang}
\affiliation{%
 \institution{Beijing Institute of Technology}
 \city{Beijing}
 \country{China}
}
\email{wanggrbit@126.com}

%\author{Aparna Patel}
%\affiliation{%
% \institution{Rajiv Gandhi University}
% \streetaddress{Rono-Hills}
% \city{Doimukh}
% \state{Arunachal Pradesh}
% \country{India}}

%\author{Huifen Chan}
%\affiliation{%
 % \institution{Tsinghua University}
  %\streetaddress{30 Shuangqing Rd}
  %\city{Haidian Qu}
  %\state{Beijing Shi}
  %\country{China}}

% \author{Anonymous Authors}

% \affiliation{ ~\\}

% \affiliation{Paper ID: 1307}

% \affiliation{ ~\\}

% \affiliation{ ~\\}

%%
%% By default, the full list of authors will be used in the page
%% headers. Often, this list is too long, and will overlap
%% other information printed in the page headers. This command allows
%% the author to define a more concise list
%% of authors' names for this purpose.
\renewcommand{\shortauthors}{Kaituo Feng et al.}

%%
%% The abstract is a short summary of the work to be presented in the
%% article.
\begin{abstract}
	Knowledge distillation (KD) has demonstrated its effectiveness to boost the performance of graph neural networks (GNNs), where its goal is to distill knowledge from a deeper teacher GNN into a shallower student GNN. However, it is actually difficult to train a satisfactory teacher GNN due to the well-known over-parametrized and over-smoothing issues, leading to invalid knowledge transfer  in practical applications. In this paper, we propose the first \textbf{Free}-direction \textbf{K}nowledge \textbf{D}istillation framework via  Reinforcement learning for GNNs, called \textbf{FreeKD}, which is no longer required to provide a deeper well-optimized teacher GNN. The core idea of our work is to collaboratively build two  shallower GNNs in an effort to exchange knowledge between them via reinforcement learning in a hierarchical way.
    As we observe that one typical GNN model often has better and worse performances at different nodes during training, we devise a dynamic and free-direction knowledge transfer strategy that consists of two levels of actions: 
% 	Based on this observation, we devise a node judge module via reinforcement learning to \emph{dynamically} determine the directions of knowledge transfer between the corresponding nodes of two networks in each training iteration, so as to capture the complementary information.  Finally, we attempt to propagate not only the soft label of the node, but also its neighborhood relations which are generated by an agent-selected part of the neighborhood nodes.
    %Based on this observation, we devise a two-level knowledge judge module via reinforcement learning to capture the complementary information. 
    1) node-level action determines the directions of knowledge transfer between the corresponding nodes of two networks; and then 2) structure-level action determines which of the local structures generated by the node-level actions to be propagated.
	In essence, our FreeKD is a general and principled framework which can be naturally compatible with  GNNs of different architectures.
	Extensive experiments on five benchmark datasets demonstrate our FreeKD  outperforms two base GNNs in a large margin, and shows its efficacy to various GNNs.
	More surprisingly, our FreeKD has comparable or even better performance than traditional KD algorithms that distill knowledge from a deeper and stronger teacher GNN.
 
\end{abstract}

%%
%% The code below is generated by the tool at http://dl.acm.org/ccs.cfm.
%% Please copy and paste the code instead of the example below.
%%

% \ccsdesc[300]{Networks~Network algorithms}

%%
%% Keywords. The author(s) should pick words that accurately describe
%% the work being presented. Separate the keywords with commas.

%% A "teaser" image appears between the author and affiliation
%% information and the body of the document, and typically spans the
%% page.

% \begin{teaserfigure}
%   \includegraphics[width=\textwidth]{sampleteaser}
%   \caption{Seattle Mariners at Spring Training, 2010.}
%   \Description{Enjoying the baseball game from the third-base
%   seats. Ichiro Suzuki preparing to bat.}
%   \label{fig:teaser}
% \end{teaserfigure}

%%
%% This command processes the author and affiliation and title
%% information and builds the first part of the formatted document.
\maketitle

% \begin{figure}[htbp]
%   \centering
%   \includegraphics[width=0.9\linewidth]{figure/t}
%   \caption {(a) Conventional methods (b) Our method}
%   \Description{A woman and a girl in white dresses sit in an open car.}
% \end{figure}

\section{Introduction}
Graph data is becoming increasingly prevalent and ubiquitous with the rapid development of the Internet, such as social networks \cite{hamilton2017inductive},  citation networks \cite{sen2008collective}, etc. To better handle graph-structured data, graph neural networks (GNNs) provide an effective means to learn node embeddings by aggregating feature information of neighborhood nodes \cite{velivckovic2017graph}. 
Because of the powerful ability in modeling relations of data, various graph neural networks have been proposed in the past decade \cite{kipf2016semi,hamilton2017inductive,velivckovic2017graph,chiang2019cluster,pei2020geom}. The representative works include GraphSAGE \cite{hamilton2017inductive}, GAT \cite{velivckovic2017graph}, GCN \cite{kipf2016semi}, etc.

%widely used for solving various tasks, including recommendation \cite{fan2019graph}, molecular property prediction   \cite{hao2020asgn}, text classification \cite{yao2019graph}, link prediction \cite{zhang2018link}, etc.

Recently, some researchers extend an interesting learning scheme, called knowledge distillation (KD) , into GNNs to further improve the performance of  GNNs \cite{yang2020distilling,yang2021extract,deng2021graph}.  
The basic idea among these methods is to optimize a shallower student  GNN model by distilling knowledge from a deeper teacher  GNN model. 
For instance, LSP \cite{yang2020distilling} proposed a local structure preserving module to transfer the topological structure information of a  GNN teacher model. 
% CPF \cite{yang2021extract} distilled the knowledge of a GNN teacher model by utilizing a parameterized label propagation to absorb structure-based knowledge and using MLP to encode the information of node attributes.
The work in \cite{yan2020tinygnn} proposed a light GNN architecture, called TinyGNN, and attempted to distill knowledge from a deep GNN teacher model to the light GNN model. 
GFKD \cite{deng2021graph} designed a data-free knowledge distillation strategy for GNNs, enabling to transfer knowledge from a GNN teacher model by generating fake graphs.

The above methods follow the same teacher-student architecture as the traditional knowledge distillation methods \cite{bucilu2006model,hinton2015distilling}, and resort to a deeper well-optimized teacher GNN for distilling knowledge. However, when applying such an architecture to GNNs, it often suffers from the following limitations:
first, it is often difficult and inefficient to train a satisfactory teacher GNN. As we know, the existing over-parameterized and  over-smoothing issues often degrade the performance of the deeper GNN model. Moreover,  training a deeper well-optimized model usually needs plenty of data and high computational costs. %Meanwhile, the over-parameterized GNN model increases the risk of overfitting \cite{li2018deeper} and the large number of layers in GNN often leads to the issue of over-smoothing \cite{rong2019dropedge}, degrading the performance of the teacher ;
Second, according to \cite{yan2020tinygnn,mirzadeh2020improved,yuan2021reinforced}, we know that a stronger teacher model may not necessarily lead to a better student model. This may be because the mismatching of the representation capacities between a teacher model and a student model makes the student model hard to mimic the outputs of a too strong teacher model. Thus, it is  difficult to find an optimal teacher GNN for a student GNN in practical applications. 
Considering that  many powerful GNN models  have been proposed in the past decade \cite{wu2020comprehensive}, this gives rise to one intuitive  thought: \emph{whether we can explore a new knowledge distillation architecture to boost the performance of GNNs, avoiding the obstacle involved by training a deeper well-optimized teacher GNN?}

\begin{figure}
  \centering
  \includegraphics[width=0.9\linewidth]{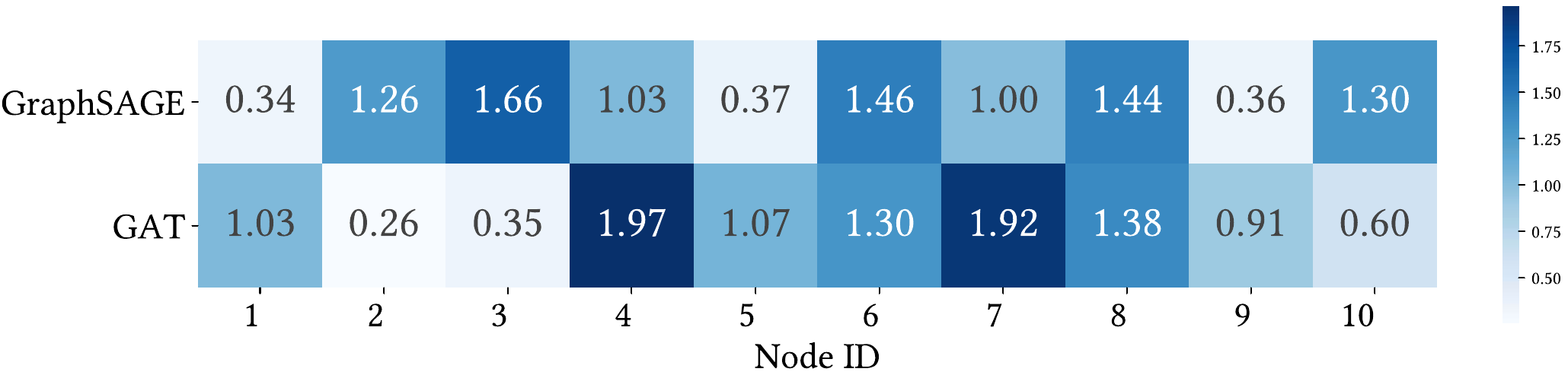}
  \caption {Cross entropy losses for nodes with ID  from $1$ to $10$ on the Cora dataset obtained by two typical GNN models, GraphSAGE \cite{hamilton2017inductive} and GAT \cite{velivckovic2017graph}, after training 20 epochs. The value in each block denotes the corresponding loss.}
  \Description{}
  \label{cross}
  \vspace{-0.15in}
\end{figure}

In light of these, we propose a new knowledge distillation framework, \textbf{Free}-direction \textbf{K}nowledge \textbf{D}istillation based on Reinforcement learning tailored for GNNs, called \textbf{FreeKD}. Rather than requiring a deeper well-optimized teacher GNN for unidirectional knowledge transfer, we collaboratively learn two  shallower  GNNs in an effort to distill knowledge from each other via reinforcement learning in a hierarchical way. This idea stems from our observation that one typical GNN model often has better and worse performances at different nodes during training. 
As shown in Figure \ref{cross}, GraphSAGE \cite{hamilton2017inductive} has lower cross entropy losses at nodes with ID$=\{1,4,5,7,9\}$, while GAT \cite{velivckovic2017graph} has better performances at the rest nodes. 
% Based on this observation, we design a reinforcement learning based node judge strategy to dynamically distinguish which model should be used to distill knowledge for each node in each training iteration, in order to capture the complementary information.
Based on this observation, we design a free-direction knowledge distillation strategy to dynamically exchange knowledge between two shallower GNNs  to benefit from each other.
Considering that the direction of distilling knowledge for each node will have influence on the other nodes, we thus regard determining the directions for different nodes as a sequential decision making problem. Meanwhile, since the selection of the directions is a discrete problem, we can not optimize it by stochastic gradient descent based methods \cite{wang2019minimax}. Thus, we address this problem via reinforcement learning in a hierarchical way. Our hierarchical reinforcement learning algorithm consists of  two levels of actions: Level 1, called  node-level action, is used to distinguish which GNN is chosen to distill knowledge to the other GNN for each node.
After determining the direction of knowledge transfer for each node, we expect to propagate not only the soft label of the node, but also its neighborhood relations.
Thus level 2, called  structure-level action, decides which of the local structures generated by our node-level actions to be propagated.
One may argue that we could directly use the loss, e.g., cross entropy, to decide the directions of  node-level knowledge distillation.
However, this heuristic strategy only considers the performance of the node itself, but neglects its influence on other nodes, thus might lead to a sub-optimal solution.
Our experimental results also verify  our reinforcement learning based strategy significantly outperforms the above heuristic one.
% After determining the direction of knowledge transfer, we expect to propagate not only the soft label of the node, but also its neighborhood relations which are generated by our node judge module. 
%In the two GNN models, each model  plays the role of either teacher or student at different nodes and {\color{blue} structures}, enabling  the two models to  flexibly learn from each other. 
    
The contributions of this paper can be summarized as:
\begin{itemize}
\item We propose a new knowledge distillation architecture for GNNs, avoiding requiring a deeper well-optimized teacher model for distilling knowledge. The proposed framework is general and principled, which can be naturally compatible with GNNs of different architectures.
% \item We devise an distillation mechanism via reinforcement learning, which can dynamically determine the direction of knowledge transfer at each node. In addition, we simultaneously transfer knowledge from both node-level and structure-level aspects to boost the performance of each model.

\item We devise a  free-direction knowledge distillation strategy via a hierarchical reinforcement learning  algorithm, which can dynamically manage the directions of knowledge transfer from both node-level and structure-level aspects.

\item Extensive experiments on five benchmark datasets demonstrate the proposed framework promotes the performance of two shallower GNNs in a large margin, and is valid to various GNNs. 	More surprisingly, the performance of our FreeKD is comparable to or even better than traditional KD algorithms distilling knowledge from a deeper and stronger teacher GNN.
\end{itemize}

\begin{figure*}[htbp]
  \centering
  \includegraphics[width=0.92\linewidth]{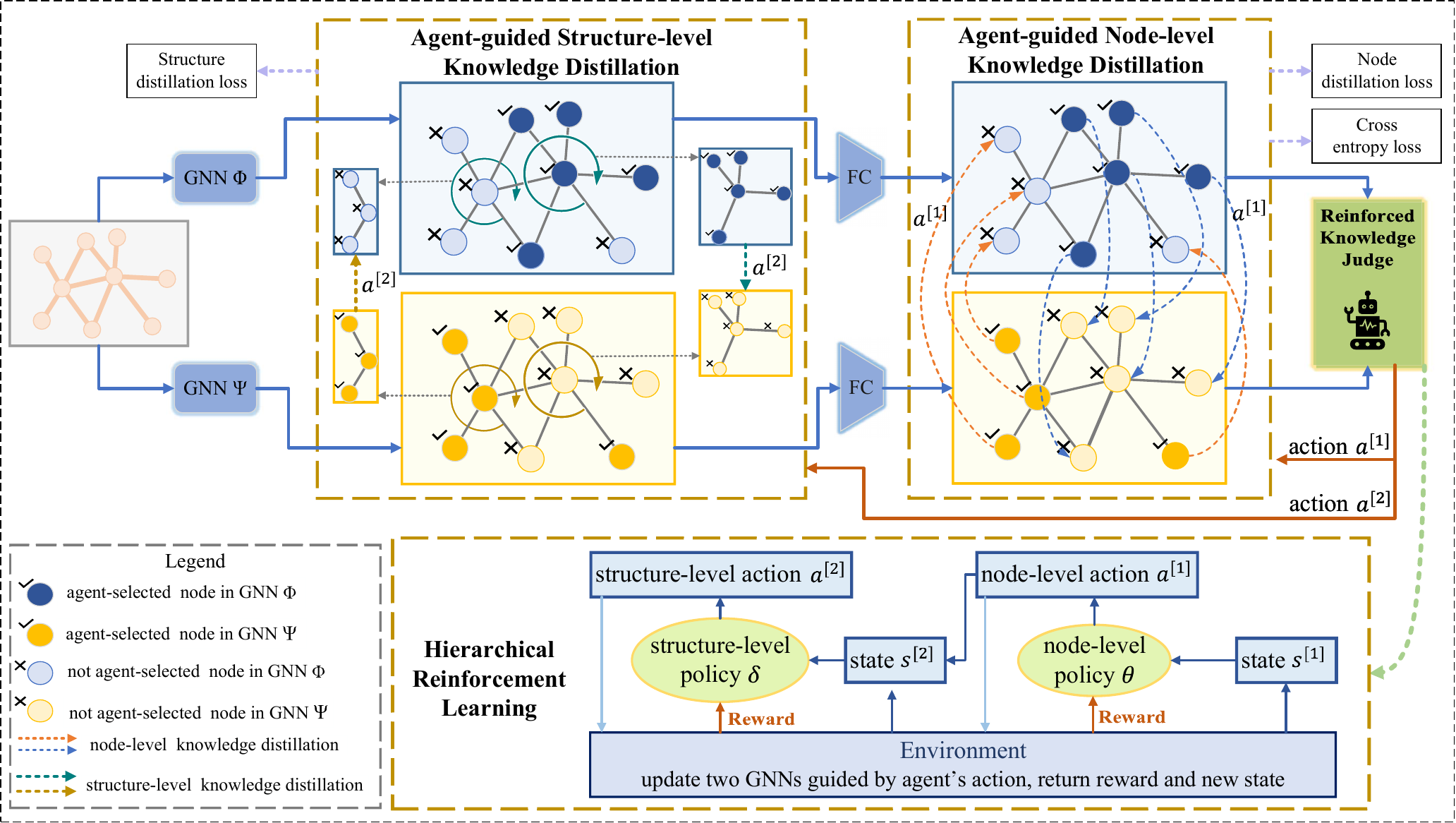}
  \caption{
    % An illustration of the BKDR framework. BKDR is mainly made up of three modules: 1) reinforced node judge module based on reinforcement learning is designed to dynamically decide the directions of knowledge distillation; 2) An agent-guided node knowledge transfer module is proposed to transfer the soft label between corresponding nodes in two GNNs; 3) An agent-guided structure knowledge transfer module intends to transfer the neighborhood relation which is generated according to the agent's actions between two GNNs.
    An illustration of the FreeKD framework. FreeKD can manage the knowledge distillation directions between two GNN models via a  hierarchical reinforcement learning that contains two-level actions. The first level of actions are designed to determine the distillation direction for each node, in order to propagate the node's soft label. And then the second level of actions are used to decide which of   the local structures generated based on node-level actions to be propagated.
  }
  \Description{Main architecture of FreeKD}
  \label{main}
\end{figure*}

\section{Related Work}
This work is related to graph neural networks, graph-based knowledge distillation, and reinforcement learning. 

\subsection{Graph Neural Networks}

Graph neural networks have achieved promising results in processing graph data, whose goal is to learn node embeddings by aggregating nodes' neighbor information.
    In recent years, lots of GNNs have been proposed \cite{kipf2016semi,hamilton2017inductive,velivckovic2017graph}. For instance, GCN \cite{kipf2016semi} designed a convolutional neural network architecture for graph data.
    GraphSAGE \cite{hamilton2017inductive} proposed an efficient sample strategy  to aggregate neighbor nodes.
    % , and can perform inductive learning in large-scale graphs with limited computational resources. 
    GAT \cite{velivckovic2017graph} applied a self-attention mechanism to GNN to assign different weights to different neighbors. 
%   The work in \cite{xu2018powerful} analyzed the upper bound  of the representation ability of GNN is the Weisfeiler-Lehman isomorphism test \cite{weisfeiler1968reduction}, and built a GNN model that could reach to this upper bound. 
   SGC \cite{wu2019simplifying} simplified GCN by removing nonlinearities and weight matrices between consecutive convolutional layers.
   ROD \cite{zhang2021rod} proposed an ensemble learning based GNN model to fuse the knowledge in multiple hops.
    APPNP \cite{klicpera2018predict} analyzed the relationship between GCN and PageRank \cite{page1999pagerank}, and proposed a propagation model combined with a personalized PageRank.
    Cluster-GCN \cite{chiang2019cluster} built an efficient model based on graph clustering \cite{schaeffer2007graph}. 
    Being orthogonal to the above approaches developing different powerful  GNN models, we concentrate on developing a new knowledge distillation framework     on the basis of various GNNs.

\subsection{Knowledge Distillation for GNNs}

    Knowledge distillation (KD) has been widely studied in computer vision \cite{chen2017learning,liu2018multi}, natural language processing \cite{kim2016sequence,liu2019improving}, etc. 
    Recently, a few KD methods have proposed for GNNs. 
    LSP \cite{yang2020distilling} transferred the topological structure knowledge from a pre-trained deeper teacher GNN to a shallower student GNN.
    CPF \cite{yang2021extract} designed a student architecture that is a combination of a parameterized label propagation and MLP layers.
    GFKD \cite{deng2021graph} proposed a method to generate fake graphs and distilled knowledge from a teacher GNN model without any training data involved.
    The authors in \cite{yan2020tinygnn} designed an efficient GNN model by utilizing the information from peer nodes to model the local structure explicitly and distilling the neighbor structure information from a deeper GNN implicitly.
    The work in \cite{chen2020self} studied a self-distillation framework, and proposed an adaptive discrepancy retaining  regularizer to empower the transferability of knowledge.
        % GNN-SD \cite{chen2020self} proposed to distill knowledge from the shallow layers to the deep layers in one GNN model.
    RDD \cite{zhang2020reliable} was a semi-supervised knowledge distillation method for GNNs. It online learnt a complicated teacher GNN model by ensemble learning, and distilled  knowledge from the generated teacher model into the student model. 
    Different from them, we focus on studying a new free-direction knowledge distillation architecture, with the purpose of dynamically exchanging knowledge between two shallower GNNs. 
% {\color{blue}
% \subsection{Ensemble learning}
% The basic idea of ensemble learning is to ensemble multiple peer models to derive a powerful model. The representative methods include Bagging \cite{oza2001online}, Boosting \cite{schapire2013explaining},  Stacking \cite{pavlyshenko2018using}, etc. 
% In common, most of ensemble learning methods derive a more accurate output by fusing the outputs of a huge number of peer models.
% Thus, despite the derived ensembled model has good performance, the high computational and storage cost is inevitable. 
% Our method is different from ensemble learning, we don't ensemble many GNNs and use their outputs to derive a powerful model. Instead, we use two GNNs to exchange learned knowledge for mutual improvement. 
% }

\subsection{Reinforcement Learning}
    Reinforcement learning aims at training agents to make optimal decisions by learning from interactions with the environment \cite{arulkumaran2017deep}. Reinforcement learning mainly has two genres \cite{arulkumaran2017deep}: value-based methods and policy-based methods. Value-based methods estimate the expected reward of actions \cite{mnih2015human}, while policy-based methods take actions according to the output probabilities of the agent \cite{williams1992simple}. There is also a hybrid of this two genres,  called the actor-critic architecture \cite{haarnoja2018soft}. The actor-critic architecture utilizes the value-based method as a value function to estimate the expected reward, and employs the policy-based method as a policy search strategy to take actions. 
    Until now,  reinforcement learning has been taken as a popular tool to solve various tasks, such as recommendation systems \cite{zheng2018drn}, anomaly detection \cite{oh2019sequential}, multi-label learning \cite{chen2018recurrent}, etc.
    In this paper, we explore  reinforcement learning for graph data based knowledge distillation.

\section{METHODOLOGY}
    In this section, we elaborate the details of our FreeKD framework that is shown in Figure \ref{main}.
    % As shown in Figure \ref{main}, our framework mainly consists of three modules: A  node judge module aims at dynamically deciding the directions of knowledge distillation via reinforcement learning to capture the complementary information for each node;
    % After determining the direction for each node, an agent-guided node knowledge transfer module attempts to propagate the soft label from one node of one network to the corresponding node of the other network; An agent-guided structure knowledge transfer module aims to transfer the neighborhood relations of the nodes which are obtained based on our node judge module.
    Before introducing it, we first give some notations and  preliminaries.

\subsection{Preliminaries}
	Let $\mathbf{G}=(\mathbf{V},\mathbf{E},\mathbf{X})$ denote a graph, where $\mathbf{V}$ is the set of nodes and $\mathbf{E}$ is the set of edge. $\ \mathbf{X}\in\mathbb{R}^{N\times d}$ is the feature matrix of nodes, where $N$ is the number of nodes and $d$ is the dimension of node features.  Let $\mathbf{x}_i$ be the feature representation of node $i$ and $y_i$ be its class label.
	The neighborhood set of node $i$ is ${\mathcal{N}(i)}=\{\ j\in \mathbf{V}\ |\ (i,j)\in \mathbf{E}\ \}$.
	%The node classification task aims to map every node to a class label. 
	Currently, graph neural networks (GNNs) have become one of the most popular models for handling graph data.
	 GNNs can learn the embedding $\mathbf{h}_i^{(l)}$  for node $i$ at the $l$-th layer  by the following formula:
\begin{align}
  \mathbf{h}_i^{(l)}=AGGREGATOR(\mathbf{h}_i^{(l-1)},\{\mathbf{h}_j^{(l-1)}|\ j\in{\mathcal{N}(i)}\},{\ \mathbf{W}^{(l)}}),
\end{align}
where $AGGREGATOR$ is an aggregation function, and it can be defined in many forms, e.g., mean aggregator \cite{hamilton2017inductive}.  $\mathbf{W}^{(l)}$ is the learnt parameters in the  $l$-th layer of the network. The initial feature of each node $i$ can be used as the input of the first layer, i.e., $\mathbf{h}_i^{(0)}=\mathbf{x}_i$.

%After learning the node representations, several  fully-connected layers can serve as a projection head to map the node representations to the probability distributions of the classes. Finally, the cross entropy loss is often used for optimizing the network.

% The cross entropy loss is often used for optimizing the network,  defined as:

% \begin{equation}
%     L_{CE}=-\sum_{i=1}^{N}\sum_{j=1}^{M}{I(y_i,j)log(p^j_{i})},
% \end{equation}
% where $M$ is the number of classes, and $p^j_{i}$ denotes the predicted probability of node $i$ belonging to class $j$. $I(y_i,j)$ is the indicator function defined as:
% \begin{equation}
% I(y_i,j)=\begin{cases}
% 0,& \text{$y_i\neq j$}\\
% 1,& \text{$y_i=j$}.
% \end{cases}
% \end{equation}

Being orthogonal to those works developing various GNN models, our goal is to explore a new knowledge distillation framework for promoting the performance of GNNs, while addressing the issue involved because of producing a deeper teacher GNN model in the existing KD methods. 
% Next, we will first introduce the node judge module which can dynamically determine the directions of knowledge distillation via reinforcement learning for each node.
% {\color{blue}
% Next, we will first introduce our two-level reinforced knowledge judge module which can dynamically determine the knowledge to transfer between two GNNs.
% }

\subsection{Overview of Framework}
As shown in Figure \ref{cross}, we observe typical GNN models often have different performances at different nodes during training.
% , i.e., one GNN obtains better performances at certain nodes but the other is superior at other nodes. 
Based on this observation, we intend to dynamically exchange useful knowledge between two shallower GNNs, so as to benefit from each other. 
However, a challenging problem is attendant upon that: how to decide the directions of knowledge distillation for different nodes during training. 
To address this, we propose to manage the directions of knowledge distillation via reinforcement learning, where we regard the directions of knowledge transfer for different nodes as a sequential decision making problem \cite{pednault2002sequential}. Consequently, we propose a free-direction knowledge distillation framework via a hierarchical reinforcement learning, as shown in Figure \ref{main}.
In our framework, the hierarchical reinforcement learning can be taken as a reinforced knowledge judge that consists of two levels of actions: 1) Level 1, called node-level action, is used to decide the distillation direction of each node for propagating the soft label; 2) Level 2, called structure-level action, is used to determine which of  the local structures generated via node-level actions to be propagated.

 Specifically, the reinforced knowledge judge (we call it agent for convenience) interacts with the environment constructed by two GNN models in each iteration, as in Figure \ref{main}.
	It receives the soft labels and cross entropy losses for a batch of nodes, and regards them as its node-level states. 
	The agent then samples sequential node-level actions for nodes according to a learned policy network, where each action decides the direction of knowledge distillation for propagating node-level knowledge. 
	Then, the agent receives the structure-level states and produces structure-level actions to decide which of the local structures generated on the basis of node-level actions to be propagated.
	After that, the two GNN models are trained based to the agent's actions with a new loss function.
	Finally, the agent calculates the reward for each action to train the policy network, where the agent's target is to maximize the expected reward.
	This process is repeatedly iterated until convergence.
% 	Note that our method is different from  traditional reinforcement learning \cite{arulkumaran2017deep}, which samples one action based on one state. Similar to \cite{yuan2021reinforced,wang2019minimax}, we sample a batch of actions based on a batch of states, and received the delayed reward after two GNNs being updated.

	We first  give some notations for convenient presentation, before introducing how to distill both node-level and structure-level knowledge.
	Let $\Phi$ and $\Psi$ denote two GNN models, respectively.  $\mathbf{h}_i^\Phi$ and $\mathbf{h}_i^\Psi$ denote the learnt representations of node $i$ obtained by $\Phi$ and $\Psi$, respectively.
	Let $\mathbf{p}_i^\Phi$ and $\mathbf{p}_i^\Psi$ be the predicted probabilities of the two GNN models for node $i$ respectively. We regard them as the soft labels.
	In addition,  $L_{CE}^\Phi(i)$ and $L_{CE}^\Psi(i)$ denote the cross entropy losses of node $i$ in $\Phi$ and $\Psi$, respectively.

\subsection{Agent-guided Node-level Knowledge Distillation}
In this section, we introduce our reinforcement learning based strategy  to dynamically distill node-level knowledge  between two GNN models. 

\subsubsection{Node-level State}  We concatenate the following features as the node-level state vector $s_i^{[1]}$ for node $i$: 

(1) Soft label vector of node $i$ in GNN $\Phi$.

(2) Cross entropy loss of node $i$ in GNN $\Phi$.

(3) Soft label vector of node $i$ in GNN $\Psi$.

(4) Cross entropy loss of node $i$ in GNN $\Psi$.

	The first two kinds of features are based on the intuition that the cross entropy loss and soft label can quantify the useful knowledge for node $i$ in GNN $\Phi$ to some extent. 
	The last two kinds of features have the same function for $\Psi$.
	Since these features can measure the knowledge each node contains to some extent, we use them as the feature of the node-level state for predicting the node-level actions.
 
	Formally, the state $s_i^{[1]}$ for the node $i$ is expressed as:
\begin{equation}\label{n-state}
    \mathbf{s}_i^{[1]}=CONCAT(\mathbf{p}_i^\Phi,L_{CE}^\Phi(i),\mathbf{p}_i^\Psi,L_{CE}^\Psi(i)),
\end{equation}
where $CONCAT$ is the concatenation operation.

\subsubsection{Node-level Action} 
% 	The agent makes a decision by outputting the action $a_i$ which is 1 or 0. 
	The node-level action $a_i^{[1]}\in\{0,1\}$ decides the direction of knowledge distillation for node $i$.
	$a_i^{[1]}=0$ means transferring knowledge from GNN $\Phi$ to GNN $\Psi$ at node $i$, while $a_i^{[1]}=1$ means the distillation direction from $\Psi$ to  $\Phi$. 
	If $a_i^{[1]}=0$, we define node $i$ in $\Phi$ as \emph{agent-selected node}, otherwise,  we define node $i$ in $\Psi$ as \emph{agent-selected node}.
	The actions are sampled from the probability distributions produced by a node-level policy function $\pi_\mathbf{\theta}$, where $\mathbf{\theta}$ is the trainable parameters in the policy network and $\pi_\mathbf{\theta}\left(s_i^{[1]},a_i^{[1]} \right)$ means the probability to take action $a_i^{[1]}$ over the state $s_i^{[1]}$. In this paper, we adopt a three-layer MLP with the $tanh$ activation function as our node-level policy network.

\begin{figure}
\centering
\subfigure[Node-level distillation]{
\begin{minipage}[t]{0.49\linewidth}
\centering
\includegraphics[width=1.46in]{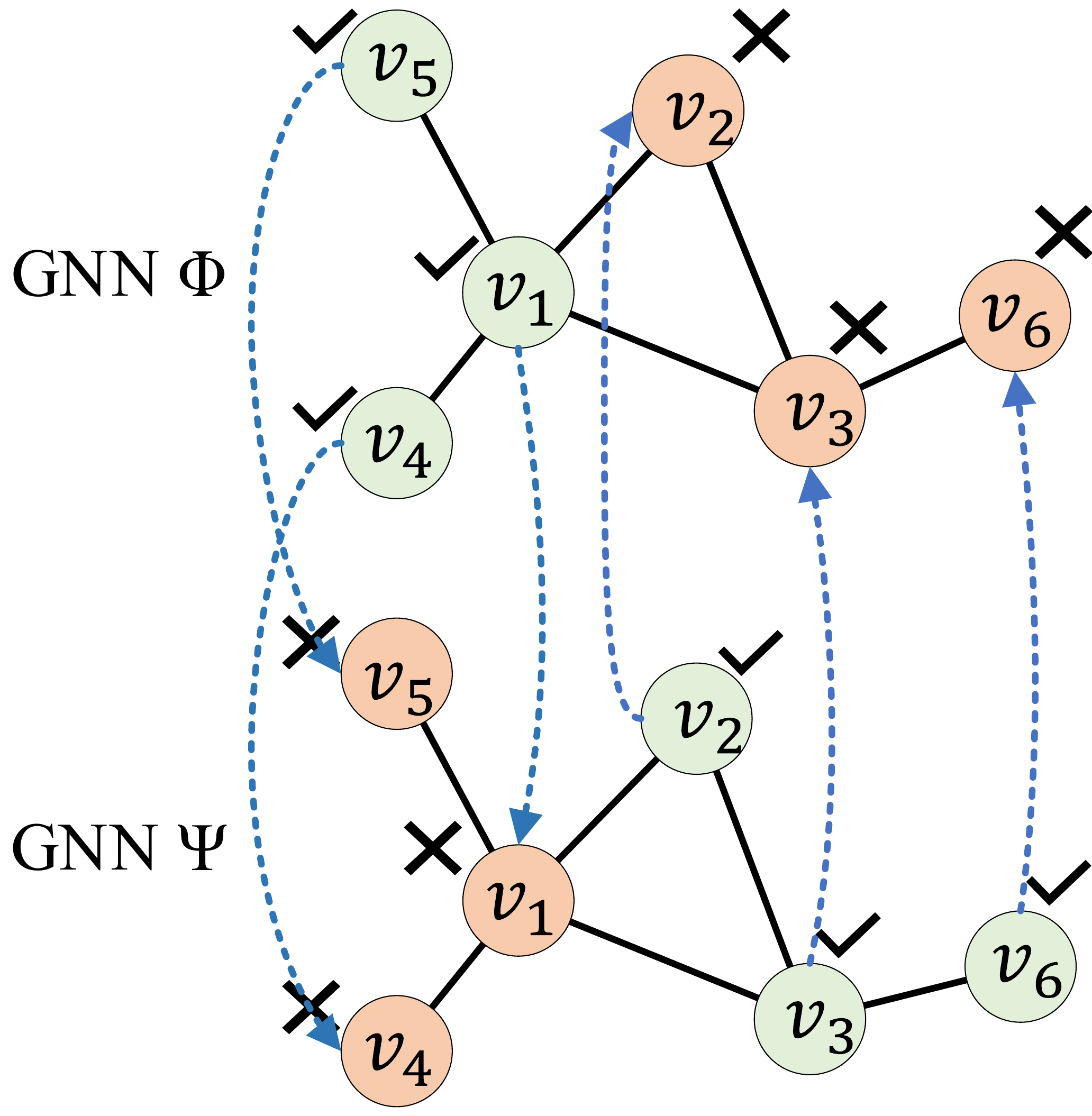}
\end{minipage}%
}%
\subfigure[Structure-level distillation]{
\begin{minipage}[t]{0.49\linewidth}
\centering
\includegraphics[width=1.2in]{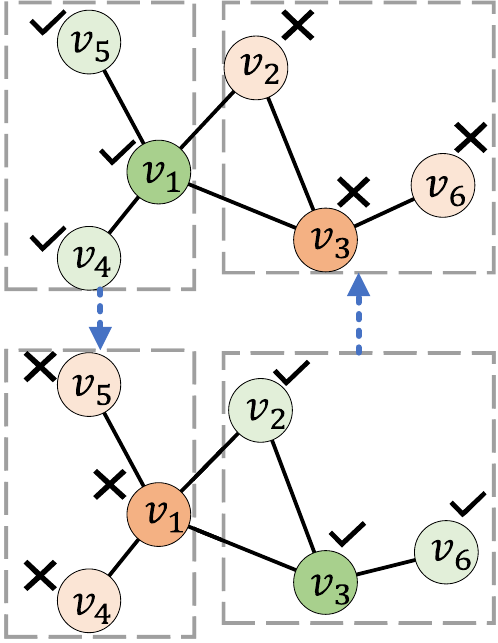}
\end{minipage}%
}%
\centering
\vspace{-0.2in}
\caption{Illustration of both node-level and structure-level knowledge distillation.}
\label{transfer}
\vspace{-0.15in}
\end{figure}

\subsubsection{Node-level Knowledge Distillation} 
    After determining the direction of knowledge distillation for each node, the two GNN models can exchange beneficial node-level knowledge. We take Figure \ref{transfer}(a) as an example to illustrate our idea. In Figure \ref{transfer}(a), the agent-selected nodes $\{v_1, v_4, v_5\}$ in GNN $\Phi$ will serve as the distilled nodes to transfer knowledge to the nodes  $\{v_1, v_4, v_5\}$  in GNN $\Psi$. In the meantime, the agent-selected nodes $\{v_2, v_3, v_6\}$ in $\Psi$ will be used as the distilled nodes to distill knowledge for the nodes $\{v_2, v_3, v_6\}$ in $\Phi$. 
    In order to transfer node-level knowledge, we utilize the KL divergence to measure the distance between the soft labels of the same node in the two GNN models, and  propose to minimize a new loss function for each GNN model:  
\begin{align}
L_{node}^\Psi=\sum_{i=1}^{N}{(1-a_i^{[1]}) KL(\mathbf{p}_i^\Phi||\mathbf{p}_i^\Psi)}\label{ce1}\\
L_{node}^\Phi=\sum_{i=1}^{N}{a_i^{[1]}KL(\mathbf{p}_i^\Psi||\mathbf{p}_i^\Phi)}\label{ce2},
\end{align}
where the value of $a_i^{[1]}$ is 0 or 1. 
When $a_i^{[1]}=0$, we use the $KL$ divergence to make the probability distribution $\mathbf{p}_i^\Psi$ match $\mathbf{p}_i^\Phi$ as much as possible, enabling the knowledge from $\Phi$ to be transferred to $\Psi$ at node $i$, and vice versa for $a_i^{[1]}=1$.
Thus, by minimizing the two loss functions $L_{node}^\Phi$ and $L_{node}^\Psi$, we can reach the goal of dynamically exchanging useful node-level knowledge between two GNN models and thus obtaining gains from each other. 

\subsection{Agent-guided Structure-level Knowledge Distillation}

As we know, the structure information is important for graph learning. Thus, we attempt to dynamically transfer structure-level knowledge between two GNNs.
It is worth noting that we don't propagate all neighborhood information of one node as structure-level knowledge. Instead, we propagate a neighborhood subset of the node, which is comprised of agent-selected nodes. This is because we think agent-selected nodes contain more useful knowledge. 
We take Figure \ref{transfer}(b) as an example to illustrate it. $v_1$ is an agent-selected node in $\Phi$. When transferring its local structure information to $\Psi$, we only transfer the local structure composed of $\{v_1, v_4, v_5\}$.
In other words, the local structure of node $v_1$ we consider to transfer is made up of agent-selected nodes.  We call it agent-selected neighborhood set.
Moreover, considering the knowledge of the local structure in graphs is not always reliable \cite{zhang2020reliable,chen2021topology}, we design a reinforcement learning based strategy to distinguish which of the local structures to be propagated.
Next, we introduce it in detail.

\subsubsection{Structure-level State} 
We adopt the following features as the structure-level state vector $s_i^{[2]}$ for the local structure of node $i$:

(1) Node-level state of node $i$.

(2) Center similarity of node $i$'s agent-selected neighborhood set in the distilled network. 

(3) Center similarity of the same node set as node $i$'s agent-selected neighborhood set in the guided network. 

Since the node-level state contains much information for measuring the information of local structures, we use the node-level state as the first feature of structure-level state. 
As \cite{xie2020gnns} points out, the center similarity can indicate the performance of GNNs, where the center similarity measures the degree of similarity between the node and its neighbors. In other words, if center similarity  is high, the structure information should be more reliable. Thus, we also take center similarity as another feature.
Motivated by \cite{xie2020gnns}, we present a similar strategy to calculate the center similarity as:

    First, let $\mathbf{M}_i^\Phi$ and $\mathbf{M}_i^\Psi$ denote the  agent-selected neighborhood set of node $i$ in $\Phi$ and $\Psi$, respectively. Formally, 
    \begin{align}
        &\mathbf{M}_i^\Phi=\{v\ |\ a_i^{[1]}=0\ , a_v^{[1]}=0, and\ \left(i,v\right)\in \mathbf{E}\}\\
        &\mathbf{M}_i^\Psi=\{v\ |\ a_i^{[1]}=1\ , a_v^{[1]}=1, and\ \left(i,v\right)\in \mathbf{E}\}.
    \end{align}
Then, we calculate the center similarity as:
    \begin{equation}
    \mathbf{u}_i\!=\!
    \begin{cases}
    (\frac{1}{|\mathbf{M}_i^\Phi|}\sum\limits_{v\in \mathbf{M}_i^\Phi}{g(\mathbf{h}_i^\Phi,\mathbf{h}_v^\Phi)},
    \frac{1}{|\mathbf{M}_i^\Phi|}\sum\limits_{v\in \mathbf{M}_i^\Phi}{g(\mathbf{h}_i^\Psi,\mathbf{h}_v^\Psi)})
    ,\emph{if} \ \  \text{$a_i^{[1]}=0$}\\
    (\frac{1}{|\mathbf{M}_i^\Psi|}\sum\limits_{v\in \mathbf{M}_i^\Psi}{g(\mathbf{h}_i^\Psi,\mathbf{h}_v^\Psi)},
    \frac{1}{|\mathbf{M}_i^\Psi|}\sum\limits_{v\in \mathbf{M}_i^\Psi}{g(\mathbf{h}_i^\Phi,\mathbf{h}_v^\Phi)}),
    \emph{if} \ \  \text{$a_i^{[1]}=1$},\\
    \end{cases}\nonumber
    \end{equation}
where $g$ can be an arbitrary similarity  function. Here we use the cosine similarity function $g(\mathbf{x},\mathbf{y})=cos(\mathbf{x},\mathbf{y})$. 
$\mathbf{u}_i$ is a two-dimension vector. In order to better present what $\mathbf{u}_i$ stands for, we take $v_1$ and $v_3$ in Figure \ref{transfer}(b) as an example. 
$v_1$ is an agent-selected node in $\Phi$, i.e., $a_1^{[1]}=0$, and $v_3$ is an agent-selected node in $\Psi$, i.e., $a_3^{[1]}=1$. 
For $\mathbf{u}_1$, 
its first element $u_1^{(1)}$ is the center similarity between $v_1$ and \{$v_4$, $v_5$\} in the distilled network $\Phi$, 
while its second element $u_1^{(2)}$ is the center similarity between $v_1$ and \{$v_4$, $v_5$\} in the guided network $\Psi$.
Similarly, for $\mathbf{u}_3$, $u_3^{(1)}$ measures the center similarity between $v_3$ and \{$v_2$, $v_6$\} in  $\Psi$, and $u_3^{(2)}$ is  the center similarity between $v_3$ and \{$v_2$, $v_6$\} $\Phi$.
In a word, the first element in $\mathbf{u}_i$ measures the center similarity in the distilled network, and the second element measures the center similarity in the guided network.

    % The motivation to use the similarity between the representation of central node and its neighbors is based on homophily assumption \cite{wang2021hagen,wang2021powerful}. ( i.e. nodes with same or similar class are tend to connect to each other, connected nodes are prone to have similar representations). If the similarity value between the embeddings of node $i$ and its neighbors in one GNN is low, that could indicate this GNN is uncertain and unstable for the local structure of node $i$ to a certain degree. In this case, the structure-level information might be not reliable enough to distill. Thus, the last two features could help to judge whether to transfer the structure-level knowledge.

	Finally, the structure-level state $s_i^{[2]}$ for the the local structure of node $i$ is expressed as:
\begin{equation}\label{s-state}
    \mathbf{s}_i^{[2]}=
    CONCAT(\mathbf{s}_i^{[1]},\mathbf{u}_i),
\end{equation}
where $CONCAT$ is the concatenation operation.

    \subsubsection{Structure-level Action} 
    Structure-level action $a_i^{[2]}\in\{0,1\}$ is the second level action that determines which of the structure-level knowledge to be propagated. 
    If $a_i^{[2]}=1$, the agent decides to transfer the knowledge of the local structure encoded in the agent-selected neighborhood set of node $i$, otherwise it will not be transferred.
    Similar to the node-level policy network, the structure-level policy network $\pi_\mathbf{\delta}$ that produces structure-level actions is also comprised of a three-layer MLP with the $tanh$ activation function.

\subsubsection{Structure-level Knowledge Distillation}
We first introduce how to distill structure-level knowledge from $\Phi$ to $\Psi$. The method for distilling from $\Psi$ to $\Phi$ is the same. First, we define the similarity between two agent-selected nodes $i$ and $j$ by:
	\begin{align}
    \hat{s}_{ij}^\Phi=\frac{e^{g\left(\mathbf{h}_i^\Phi,\mathbf{h}_j^\Phi\right)}}{\sum_{v\in \mathbf{M}_i^\Phi} e^{g\left(\mathbf{h}_i^\Phi,\mathbf{h}_v^\Phi\right)}},\ \ 
    \hat{s}_{ij}^\Psi=\frac{e^{g\left(\mathbf{h}_i^\Psi,\mathbf{h}_j^\Psi\right)}}{\sum_{v\in \mathbf{M}_i^\Phi} e^{g\left(\mathbf{h}_i^\Psi,\mathbf{h}_v^\Psi\right)}},
\end{align}
where $g$ is the cosine similarity function.

To transfer structure-level knowledge, we propose a new loss function to be minimized as:
\begin{align}\label{struct1}
    L_{struct}^\Psi=\sum_{i=1}^{N}{(1-a_i^{[1]})a_i^{[2]} KL(\mathbf{\hat{s}}_i^\Phi||\mathbf{\hat{s}}_i^\Psi)},
\end{align}
where $\mathbf{\hat{s}}_i^\Phi=[\hat{s}_{i1}^\Phi, \cdots, \hat{s}_{iC_i^{\Phi}}^\Phi]$ and $\mathbf{\hat{s}}_i^\Psi=[\hat{s}_{i1}^\Psi, \cdots, \hat{s}_{iC_i^{\Phi}}^\Psi]$, $C_i^{\Phi}$ is the size of  $\mathbf{M}_i^\Phi$. $\mathbf{\hat{s}}_i^\Phi$ represents the distribution of the similarities between node $i$ and its agent-selected neighborhoods in $\Phi$, while $\mathbf{\hat{s}}_i^\Psi$ represents the distribution of the similarities between node $i$ and its corresponding neighborhoods in $\Psi$.
If the local structure of node $i$ is decided to transfer, we adopt the $KL$ divergence to make $\mathbf{\hat{s}}_i^\Psi$ match $\mathbf{\hat{s}}_i^\Phi$, so as to transfer structure-level knowledge.
Similarly, we can propose another new loss function for distilling knowledge from $\Psi$ to $\Phi$ as:
\begin{equation}\label{struct2}
    L_{struct}^\Phi=\sum_{i=1}^{N}{a_i^{[1]}a_i^{[2]}KL(\mathbf{\bar{s}}_i^\Psi||\mathbf{\bar{s}}_i^\Phi)},
\end{equation}
where $\mathbf{\bar{s}}_i^\Phi=[\bar{s}_{i1}^\Phi, \cdots, \bar{s}_{iC_i^{\Psi}}^\Phi]$ and $\mathbf{\bar{s}}_i^\Psi=[\bar{s}_{i1}^\Psi, \cdots, \bar{s}_{iC_i^{\Psi}}^\Psi]$, $C_i^{\Psi}$ is the size of  $\mathbf{M}_i^\Psi$. $\bar{s}_{ij}^\Phi$ and $\bar{s}_{ij}^\Psi$ are defined as:
\begin{align}
    \bar{s}_{ij}^\Phi=\frac{e^{g\left(\mathbf{h}_i^\Phi,\mathbf{h}_j^\Phi\right)}}{\sum_{v\in \mathbf{M}_i^\Psi} e^{g\left(\mathbf{h}_i^\Phi,\mathbf{h}_v^\Phi\right)}}, \ \ 
    \bar{s}_{ij}^\Psi=\frac{e^{g\left(\mathbf{h}_i^\Psi,\mathbf{h}_j^\Psi\right)}}{\sum_{v\in \mathbf{M}_i^\Psi} e^{g\left(\mathbf{h}_i^\Psi,\mathbf{h}_v^\Psi\right)}}.
\end{align}

By jointly minimizing (\ref{struct1}) and (\ref{struct2}), we can dynamically exchange structure-level knowledge  between $\Phi$ and $\Psi$.

\subsection{Optimizations}
In this section, we introduce the optimization procedure of our method.
The detailed training procedure is in Appendix \ref{procedure}. 

\subsubsection{Reward} 
Following \cite{wang2019minimax}, our actions are sampled in batch, and obtain the delayed reward after two GNNs being updated according to a batch of sequential actions.
Similar to \cite{ liang2021reinforced}, we utilize the performance of the models after being updated as the reward.
We use the negative value of the cross entropy loss to measure the performance of the models as in \cite{yuan2021reinforced, liang2021reinforced}, defined as:
 \begin{equation}\label{reward}
     R_i=-\frac{\sum\limits_{u\in \mathbf{B}}{(L_{CE}^\Phi(u){+L}_{CE}^\Psi(u))}}{\left|\mathbf{B}\right|}-\gamma\frac{\sum\limits_{v\in {\mathcal{N}_i}}{(L_{CE}^\Phi(v){+L}_{CE}^\Psi(v))}}{\left|{\mathcal{N}_i}\right|},
 \end{equation}
where $\gamma$ is a hyper-parameter.
$R_i$ is the reward for the action taken at node $i$, and $\mathbf{B}$ is a batch set of nodes from the training set.
The reward for an action  $a_i$ consists of two parts: The first part is the average performance for a batch of nodes,  measuring the global effects that the action $a_i$ brings on the GNN model; The second part is the average performance of the neighborhoods of node $i$,  in order to model the local effects of $a_i$. 

\subsubsection{Optimization for Policy Networks} 
Following previous studies about hierarchical reinforcement learning \cite{liu2021rmm}, the gradient of expected cumulative reward $\mathrm{\nabla}_{\theta,\delta}J$ could be computed as follows:
\begin{equation}\label{grad}
\mathrm{\nabla}_{\theta,\delta}J\!=\!
\frac{1}{|\mathbf{B}|}\!
\sum_{i\in\mathbf{B}\!}{\!(R_i\!-\!b_i)\mathrm{\nabla}_{\theta,\delta}\!\log(\pi_\theta(\mathbf{s}_i^{[1]}\!,a_i^{[1]})\pi_\delta(\mathbf{s}_i^{[2]}\!,a_i^{[2]})}),
\end{equation}
where $\theta$, $\delta$ is the learned parameters of the node-level policy network and structure-level policy network, respectively. Similar to \cite{lai2020policy}, to speed up convergence and reduce variance ,  we also add a baseline reward $b_i$ that is the rewards at node $i$ in the last epoch. The motivation behind this is to encourage the agent to achieve better performance than that of the last epoch. Finally, we update the parameters of policy networks by gradient ascent \cite{williams1992simple} as:
\begin{equation}\label{update}
    \theta\gets\theta+\eta\mathrm{\nabla}_\theta J,\ \   \delta\gets\delta+\eta\mathrm{\nabla}_\delta J,
\end{equation}
where $\eta$ is the learning rate for reinforcement learning.

\subsubsection{Optimization for GNNs}
We minimize the following loss functions for optimizing  $\Phi$ and $\Psi$, respectively:
\begin{equation}\label{loss1}
    L^\Phi=L_{CE}^\Phi+\mu L_{node}^\Phi+\rho L_{struct}^\Phi
\end{equation}
\begin{equation}\label{loss2}
    L^\Psi=L_{CE}^\Psi+\mu L_{node}^\Psi+\rho L_{struct}^\Psi,   
\end{equation}
where $L_{CE}^\Phi$, $L_{CE}^\Psi$ are the cross entropy losses for  $\Phi$ and  $\Psi$, respectively. $L_{node}^\Phi$ and $L_{node}^\Psi$ are two node-level knowledge distillation losses. $L_{struct}^\Phi$ and $L_{struct}^\Psi$ are two structure-level distillation losses. 
$\mu$ and $\rho$ are two trade-off parameters.
% By jointly minimizing them, we can mutually distill useful knowledge from the two GNN models, so as to benefit from each other. 

\begin{table*}
\caption{Results (\%) of the compared approaches for node classification in the transductive settings on the Cora, Chameleon, Citeseer, and Texas datasets.
The values in the brackets denote the performance improvement of our FreeKD over the corresponding baselines.
Here, we denote GraphSAGE as GSAGE for short.}
\vspace{-0.1in}
\setlength{\tabcolsep}{0.8mm}{%
\begin{tabular}{@{}ccccccccccc@{}}
\toprule
                            & \multicolumn{2}{c}{}             & \multicolumn{2}{c}{Cora}                         & \multicolumn{2}{c}{Chameleon}                     & \multicolumn{2}{c}{Citeseer}                    & \multicolumn{2}{c}{Texas}             \\ \midrule
\multicolumn{1}{c|}{Method} & \multicolumn{2}{c|}{Basic Model}     & \multicolumn{2}{c|}{F1 Score ($\uparrow$Impv.)}           & \multicolumn{2}{c|}{F1 Score ($\uparrow$Impv.)}            & \multicolumn{2}{c|}{F1 Score ($\uparrow$Impv.)}           & \multicolumn{2}{c}{F1 Score ($\uparrow$Impv.)} \\
\multicolumn{1}{c|}{}       &  $\Phi$ & \multicolumn{1}{c|}{ $\Psi$} & $\Phi$         & \multicolumn{1}{c|}{$\Psi$}         & $\Phi$         & \multicolumn{1}{c|}{$\Psi$}         & $\Phi$         & \multicolumn{1}{c|}{$\Psi$}         & $\Phi$              & $\Psi$              \\
\hline
\multicolumn{1}{c|}{GCN} & -  & \multicolumn{1}{c|}{-}     & 85.12        & \multicolumn{1}{c|}{-}             & 33.09        & \multicolumn{1}{c|}{-}             & 75.42        & \multicolumn{1}{c|}{-}             & 57.57             &       -            \\
\multicolumn{1}{c|}{GSAGE} & - & \multicolumn{1}{c|}{-}     & 85.36        & \multicolumn{1}{c|}{-}             & 48.77        & \multicolumn{1}{c|}{-}             & 76.56        & \multicolumn{1}{c|}{-}             & 76.22             &          -         \\
\multicolumn{1}{c|}{GAT} & -  & \multicolumn{1}{c|}{-}     & 85.45        & \multicolumn{1}{c|}{-}             & 40.29       & \multicolumn{1}{c|}{-}             & 75.66         & \multicolumn{1}{c|}{-}             & 57.84             &        -           \\
\hline
\multicolumn{1}{c|}{FreeKD}  & GCN  & \multicolumn{1}{c|}{GCN}  & 86.53(\textbf{$\uparrow$1.41}) & \multicolumn{1}{c|}{86.62(\textbf{$\uparrow$1.50})} & 37.61(\textbf{$\uparrow$4.52})  & \multicolumn{1}{c|}{37.70(\textbf{$\uparrow$4.61})} & 77.28(\textbf{$\uparrow$1.86}) & \multicolumn{1}{c|}{77.33(\textbf{$\uparrow$1.91})} & 60.28(\textbf{$\uparrow$2.71})      & 60.55(\textbf{$\uparrow$2.98})      \\
\multicolumn{1}{c|}{FreeKD}  & GSAGE & \multicolumn{1}{c|}{GSAGE} & 86.41(\textbf{$\uparrow$1.05}) & \multicolumn{1}{c|}{86.55(\textbf{$\uparrow$1.19})} & 49.89(\textbf{$\uparrow$1.12}) & \multicolumn{1}{c|}{49.85(\textbf{$\uparrow$1.08})} &  77.78(\textbf{$\uparrow$1.22}) & \multicolumn{1}{c|}{ 77.58(\textbf{$\uparrow$1.02})} & 78.76(\textbf{$\uparrow$2.54})      & 77.85(\textbf{$\uparrow$1.63})      \\
\multicolumn{1}{c|}{FreeKD}  & GAT  & \multicolumn{1}{c|}{GAT}  & 86.46(\textbf{$\uparrow$1.01}) & \multicolumn{1}{c|}{86.68(\textbf{$\uparrow$1.23})} & 43.96(\textbf{$\uparrow$3.67}) & \multicolumn{1}{c|}{44.42(\textbf{$\uparrow$4.13})} &  77.13(\textbf{$\uparrow$1.47}) & \multicolumn{1}{c|}{77.42(\textbf{$\uparrow$1.76})} & 61.18(\textbf{$\uparrow$3.34})      & 61.36(\textbf{$\uparrow$3.52})      \\
\multicolumn{1}{c|}{FreeKD}  & GCN  & \multicolumn{1}{c|}{GAT}  & 86.65(\textbf{$\uparrow$1.53}) & \multicolumn{1}{c|}{86.72(\textbf{$\uparrow$1.27})} & 35.58(\textbf{$\uparrow$2.49}) & \multicolumn{1}{c|}{43.79(\textbf{$\uparrow$3.53})} &  77.39(\textbf{$\uparrow$1.97}) & \multicolumn{1}{c|}{77.58(\textbf{$\uparrow$1.92})} & 61.06(\textbf{$\uparrow$3.49})      & 60.38(\textbf{$\uparrow$2.54})      \\
\multicolumn{1}{c|}{FreeKD}  & GCN  & \multicolumn{1}{c|}{GSAGE} & 86.26(\textbf{$\uparrow$1.14}) & \multicolumn{1}{c|}{86.76(\textbf{$\uparrow$1.40})} & 35.39(\textbf{$\uparrow$2.30}) & \multicolumn{1}{c|}{49.89(\textbf{$\uparrow$1.12}) } &  77.08(\textbf{$\uparrow$1.66}) & \multicolumn{1}{c|}{77.68(\textbf{$\uparrow$1.12})} & 60.61(\textbf{$\uparrow$3.04})      & 77.58(\textbf{$\uparrow$1.36})      \\
\multicolumn{1}{c|}{FreeKD}  & GAT  & \multicolumn{1}{c|}{GSAGE} & 86.67(\textbf{$\uparrow$1.22}) & \multicolumn{1}{c|}{86.84(\textbf{$\uparrow$1.48})} & 43.96(\textbf{$\uparrow$3.67})  & \multicolumn{1}{c|}{49.87(\textbf{$\uparrow$1.10})} & 77.24(\textbf{$\uparrow$1.58}) & \multicolumn{1}{c|}{77.62(\textbf{$\uparrow$1.06}) } & 62.45(\textbf{$\uparrow$4.61})      & 78.36(\textbf{$\uparrow$2.14})      \\ \bottomrule
\end{tabular}%
}
\label{trasductive}
\end{table*}

\section{EXPERIMENTS}
To verify the effectiveness of our proposed FreeKD, we perform the experiments on five benchmark datasets of different domains and on GNN models of different architectures. More implementation details are given in Appendix \ref{details}.
	
\subsection{Experimental Setups}
\subsubsection{Datasets} 
 We use five widely used benchmark datasets to evaluate our methods.
 Cora \cite{sen2008collective} and Citeseer \cite{sen2008collective} are two citation datasets where  nodes represent documents and   edges represent citation relationships. 
Chameleon \cite{rozemberczki2021multi} and Texas \cite{pei2020geom} are two web network datasets where  nodes stand for web pages and  edges show their hyperlink relationships. 
The PPI dataset \cite{hamilton2017inductive} consists of 24 protein–protein interaction graphs, corresponding to different human tissues.
In Appendix \ref{datasets}, we give more information about these datasets.
Following \cite{chen2018fastgcn} and \cite{huang2018adaptive}, we use 1000 nodes for testing, 500 nodes for validation, and the rest for training on the Cora and Citeseer datasets.
For Chameleon and Texas datasets, we randomly split nodes of each class into 60\%, 20\%, and 20\% for training, validation and testing respectively, following \cite{pei2020geom} and \cite{chen2020simple}.
For the PPI dataset, we use 20 graphs for training, 2 graphs for validation, and 2 graphs for testing, as in  \cite{chen2020simple}.
Following previous works \cite{velivckovic2017graph,pei2020geom} , we study the transductive setting on the first four datasets, and the inductive setting on the PPI dataset.
In the tasks of transductive setting, we predict the labels of the nodes observed during training, whereas in the task of inductive setting, we predict the labels of nodes in never seen graphs before.

\begin{table}
\caption{Results (\%) of the compared approaches for node classification in the inductive setting on the PPI dataset.
% The values in the brackets denote the performance improvement of our FreeKD over the corresponding baselines.
}
\vspace{-0.1in}
\setlength{\tabcolsep}{1.2mm}{%
\begin{tabular}{@{}ccccc@{}}
\toprule
                            & \multicolumn{2}{c}{}             & \multicolumn{2}{c}{PPI}               \\ \midrule
\multicolumn{1}{c|}{Method} & \multicolumn{2}{c|}{Basic Model}     & \multicolumn{2}{c}{F1 Score ($\uparrow$Impv.)} \\
\multicolumn{1}{c|}{}       & $\Phi$ & \multicolumn{1}{c|}{$\Psi$} & $\Phi$              & $\Psi$              \\
\hline
\multicolumn{1}{c|}{GSAGE} & - & \multicolumn{1}{c|}{-}     & 69.28             &       -            \\
\multicolumn{1}{c|}{GAT} & -  & \multicolumn{1}{c|}{-}     & 97.30              &        -           \\
\hline
\multicolumn{1}{c|}{FreeKD}  & GSAGE & \multicolumn{1}{c|}{GSAGE} & 71.72(\textbf{$\uparrow$2.44})      & 71.56(\textbf{$\uparrow$2.28})      \\
\multicolumn{1}{c|}{FreeKD}  & GAT  & \multicolumn{1}{c|}{GAT}  & 98.79(\textbf{$\uparrow$1.49})      & 98.73(\textbf{$\uparrow$1.43})      \\
\multicolumn{1}{c|}{FreeKD}  & GAT  & \multicolumn{1}{c|}{GSAGE} & 98.61(\textbf{$\uparrow$1.31})      & 72.39(\textbf{$\uparrow$3.11})      \\
\bottomrule
\end{tabular}%
}
\vspace{-0.1in}
\label{inductive}
\end{table}

\subsubsection{Baselines} In the experiment, we adopt three popular GNN models, GCN \cite{kipf2016semi}, GAT \cite{velivckovic2017graph}, GraphSAGE \cite{hamilton2017inductive}, as our basic models in our method.
Our framework aims to promote the performance of these GNN models. Thus, these three GNN models can be used as our baselines.
Since we propose a free-direction knowledge distillation framework, we also compare  with five typical knowledge distillation approaches proposed recently, including KD \cite{hinton2015distilling}, LSP \cite{yang2020distilling}, CPF \cite{yang2021extract}, RDD \cite{zhang2020reliable}, and GNN-SD \cite{chen2020self},  to further verify the effectiveness of our method.
Following \cite{chen2018fastgcn,hamilton2017inductive}, we use the Micro-F1 score as the evaluation measure throughout the experiment.

\subsection{Overall Evaluations on Our Method}
	In this subsection, we evaluate our method using three popular GNN models, GCN \cite{kipf2016semi}, GAT \cite{velivckovic2017graph}, and GraphSAGE \cite{hamilton2017inductive}. 
	We arbitrarily select two networks from the above three models as our basic models $\Phi$ and $\Psi$, and perform our method FreeKD
	, enabling them to learn from each other. 
	Note that we do not perform GCN on the PPI dataset, because of the inductive setting.

    %Table 2 shows the experiment results. We denote the GNN $\Phi$ as net 1 and GNN $\Psi$ and net 2. The “Single” represents the original training method which trains a single GNN model independently. The “BKDR” represent our training method which makes two GNNs learning from the good aspects of each other. We keep all the settings and parameters consistent for training independently and training by our method for fairness. And we explore BKDR can bring how much improvements to GNNs.
	
	Table \ref{trasductive} and Table \ref{inductive} report the experimental results.
	As shown in Table \ref{trasductive} and Table \ref{inductive}, our FreeKD can consistently promote the performance of the basic GNN models in a large margin on all the datasets.
	For instance, our method can achieve more than $4.5\%$ improvement by mutually learning from two GCN models on the Chameleon dataset, compared with the single GCN model. In summary, for the transductive learning tasks,  our method improves the performance by 1.01\% $\sim$ 1.97\% on the Cora and Citeseer datasets and 1.08\% $\sim$ 4.61\% on the Chameleon and Texas datasets, compared with the corresponding GNN models. 
	For the inductive learning task, our method improves the performance by 1.31\% $\sim$ 3.11\% on the PPI dataset dataset.
	In addition, we observe that two GNN models either sharing the same architecture or using different architectures can both benefit from each other by using our method, which shows the efficacy to various GNN models.

\subsection{Comparison with Knowledge Distillation}

Since our method is related to knowledge distillation, we also compare with the existing knowledge distillation methods to further verify effectiveness of our method. In this experiment, we first compare with three traditional knowledge distillation methods, KD \cite{hinton2015distilling}, LSP \cite{yang2020distilling}, CPF \cite{yang2021extract} distilling knowledge from a deeper and stronger  teacher GCNII model \cite{chen2020simple} into a shallower student  GAT model. 
The structure details of GCNII and GAT could be found in the Appendix \ref{details}.
%Note that for CPF, we use the student network proposed in the original paper \cite{yang2021extract}, which is a combination of parameterized label propagation and feature transformation MLP.
In addition, we also compare with an ensemble learning method, RDD \cite{zhang2020reliable}, where a complex teacher network is generated by ensemble learning for distilling knowledge. Finally, we take GNN-SD \cite{chen2020self} as another baseline, which distills knowledge  from shallow layers into deep layers in one GNN.
For our FreeKD, we take two GAT sharing the same structure as the basic models.

Table \ref{kd} lists the experimental results. Surprisingly, our FreeKD perform comparably or even better than the traditional knowledge distillation methods (KD, LSP, CPF) on all the datasets. This demonstrates the effectiveness of our method, as they  distill knowledge from the stronger teacher GCNII while we only mutually distill knowledge between two shallower GAT.
%do not require a large well-optimized teacher GNN for distillation. 
% An interesting point is that our method obtains better performance than KD and LSP. This illustrates that our free-directional knowledge distillation can obtain more gains to boost the performance of GNN models, compared to that distilling knowledge from one single direction.
In addition, our FreeKD consistently outperforms GNN-SD and RDD, which further illustrates the effectiveness of our proposed  FreeKD.

\begin{table}
\caption{Results (\%) of  different knowledge distillation methods. '-' means not available.
}
\vspace{-0.1in}
\setlength{\tabcolsep}{2mm}{
\begin{tabular}{c|c|c|c|c|c}
 \toprule
        & Cora                  & Chameleon              & Citeseer              & Texas                  & PPI                    \\ \hline
Teacher & \multirow{2}{*}{87.80} & \multirow{2}{*}{46.79} & \multirow{2}{*}{78.60} & \multirow{2}{*}{65.14} & \multirow{2}{*}{99.41} \\
GCNII   &                       &                        &                       &                        &                        \\ \hline
KD      & 86.13                 & 43.69                  & 77.03                 & 59.46                  & 97.81                  \\ 
LSP     & 86.25                 & 44.01                  & 77.21                 & 59.73                  & 98.25                  \\
CPF     & 86.41                 & 41.40                   & \textbf{77.80}                  & 60.81                  & -                      \\ \hline
GNN-SD  & 85.75                 & 40.79                  & 75.96                 & 58.65                  & 97.73                  \\
RDD     & 85.84                 & 41.15                  & 76.02                 & 58.92                  & 97.66                  \\
FreeKD    & \textbf{86.68}    & \textbf{44.42}  & 77.42                 & \textbf{61.36}      & \textbf{98.79}    \\ \bottomrule             
\end{tabular}%
}
\vspace{-0.2in}
\label{kd}
\end{table}

\begin{figure*}
\centering
\subfigure[$\Phi$: without noise; $\Psi$: without noise]{
\begin{minipage}[t]{0.3\linewidth}
\centering
\includegraphics[width=2in]{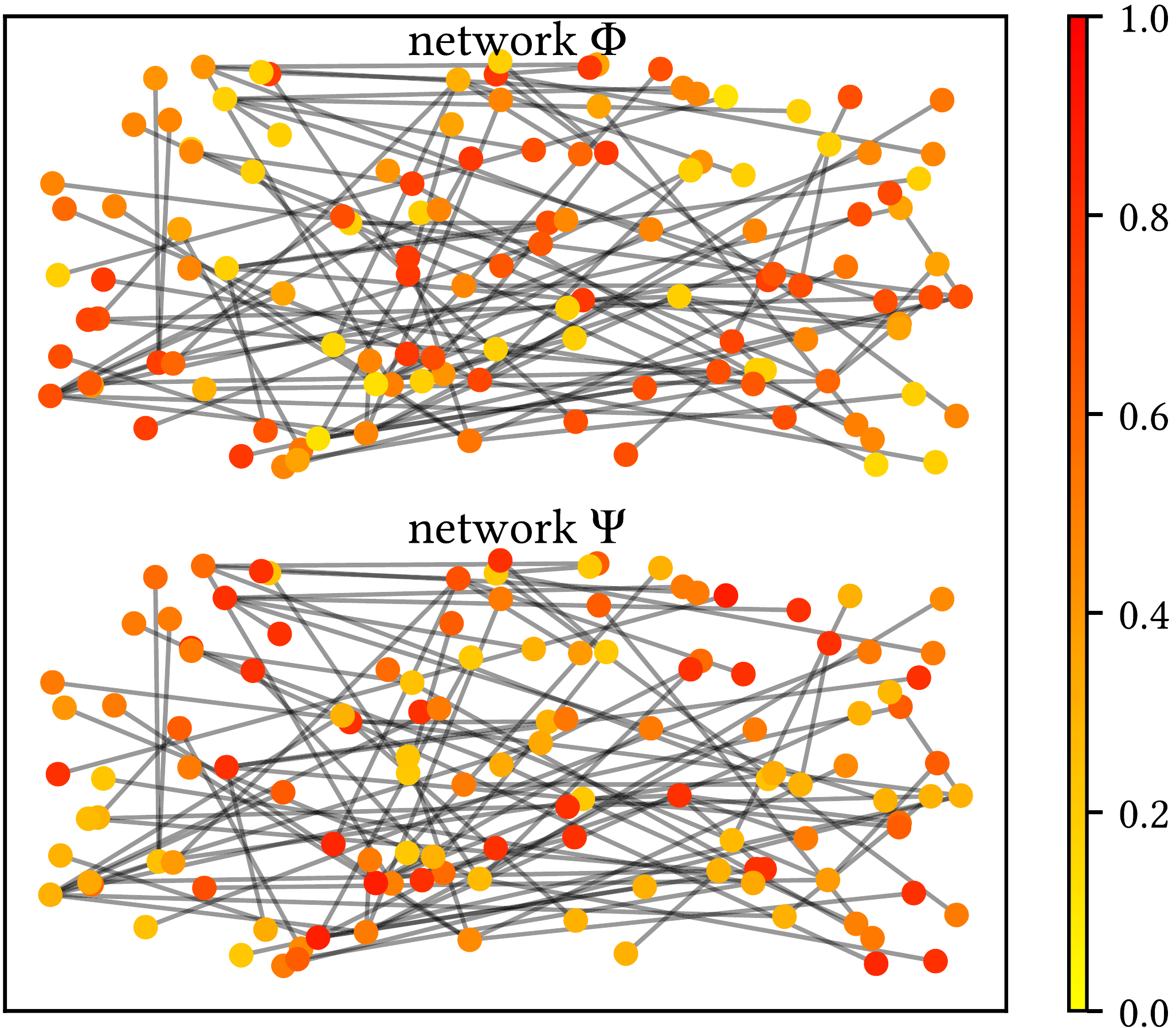}
\end{minipage}%
}%
\subfigure[$\Phi$: noise $\sigma=0.5$; $\Psi$: without noise]{
\begin{minipage}[t]{0.3\linewidth}
\centering
\includegraphics[width=2in]{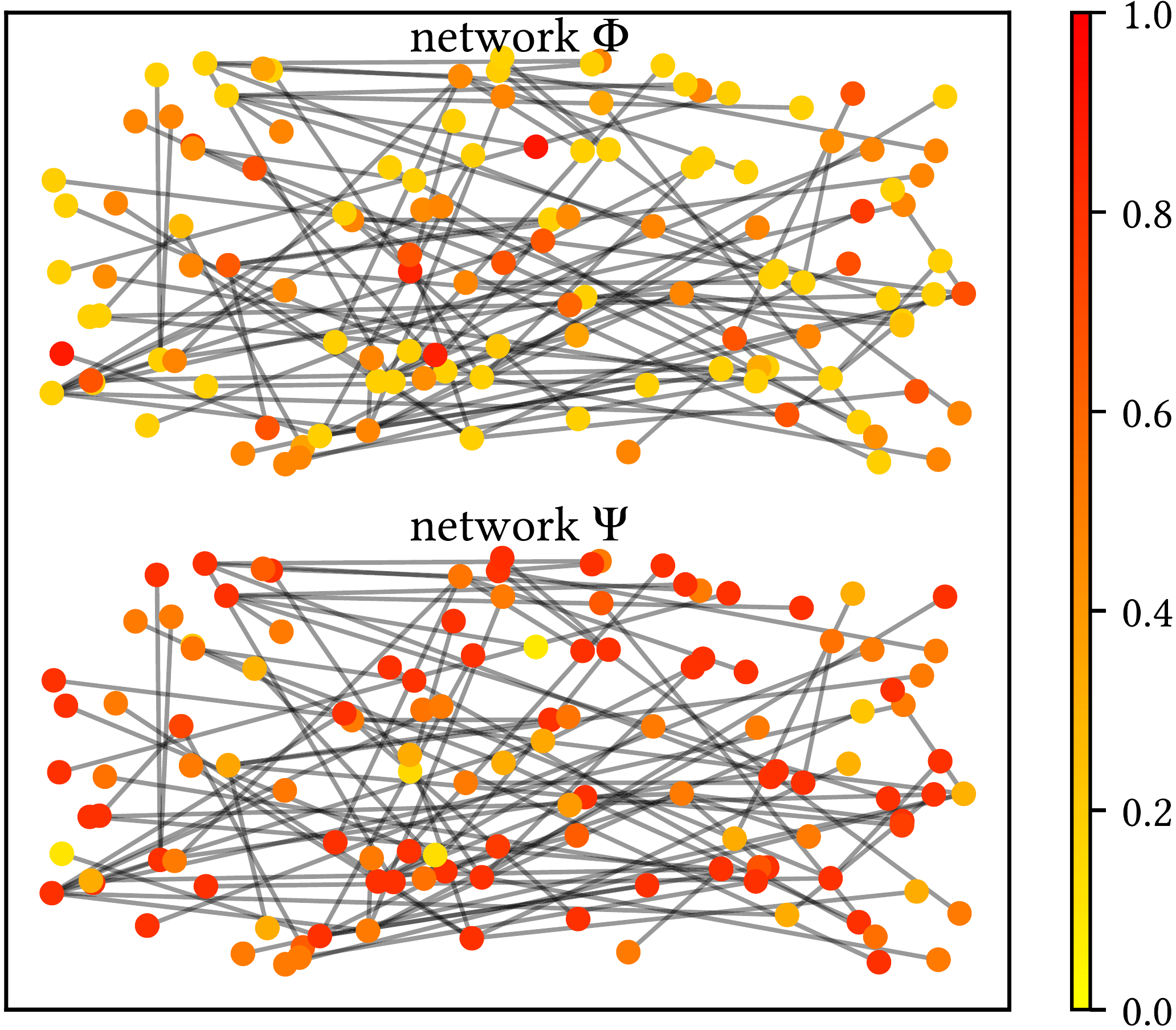}
%\caption{fig2}
\end{minipage}%
\label{vis2}
}%
\subfigure[$\Phi$: noise $\sigma=1.0$; $\Psi$: without noise]{
\begin{minipage}[t]{0.3\linewidth}
\centering
\includegraphics[width=2in]{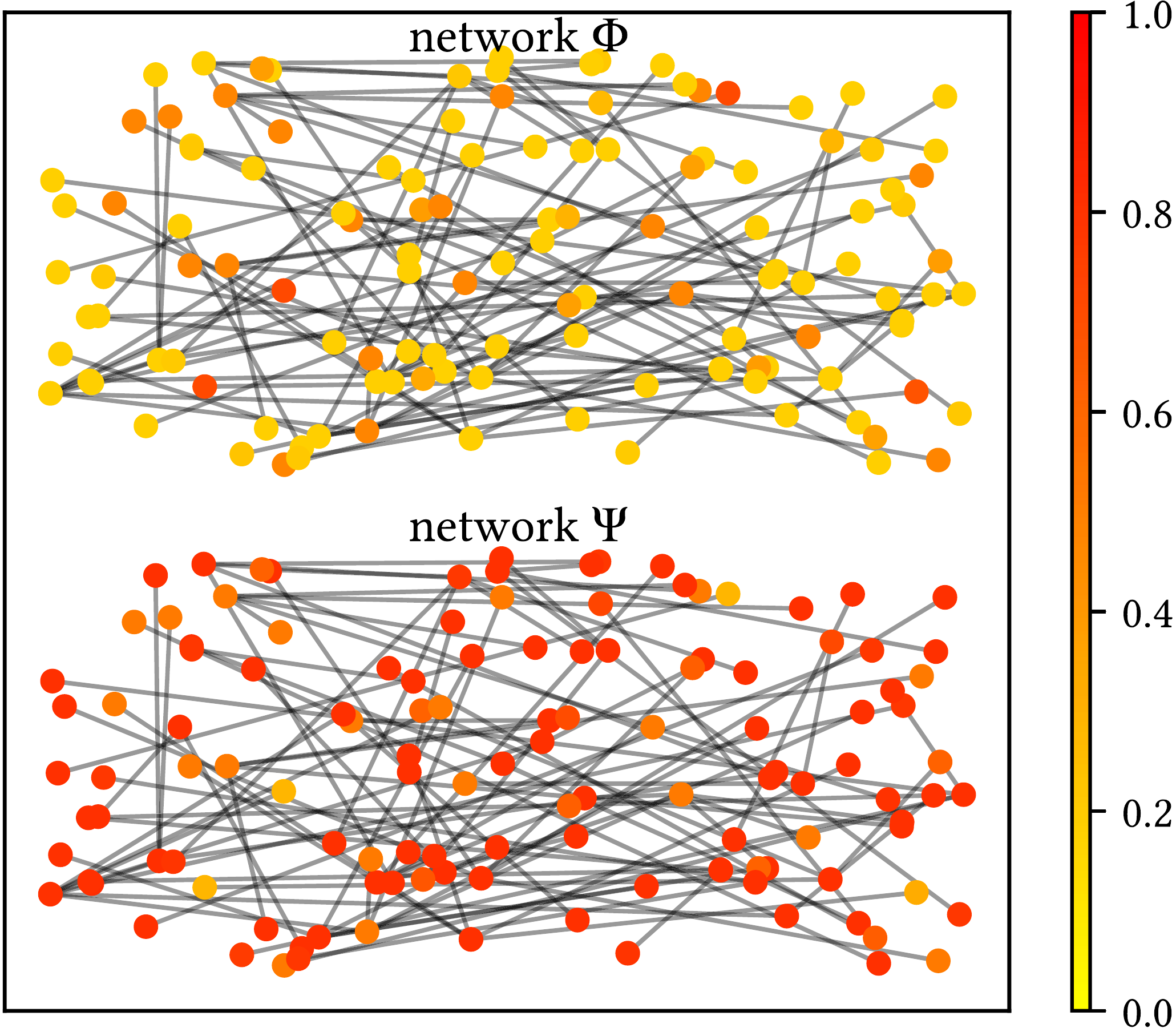}
%\caption{fig2}
\end{minipage}
\label{vis3}
}%
\centering
\vspace{-0.15in}
  \caption{The output probabilities of our node-level policy reinforced knowledge judge module by adding different degrees of noise to the network $\Phi$. 
  The redder the node in a network is, the higher the probability for the node in this network to serve as a teacher node to transfer knowledge to the corresponding node of the other network is.}
  \label{vis}
  \vspace{-0.12in}
\end{figure*}

\subsection{Ablation Study}
We perform ablation study to verify the effectiveness of the components in our method. We use GCN as the basic models $\Phi$ and $\Psi$ in our method, and conduct the experiments on two datasets of different domains, Chameleon and Cora. 
When setting $\rho=0$, this means that we only transfer the node-level knowledge. We denote it FreeKD-node for short.
To evaluate our reinforcement learning based node judge module, we design three variants:
\begin{itemize}
    \item FreeKD-w.o.-judge: our FreeKD  without using the agent.
     $\Phi$ and $\Psi$ distills knowledge for each node from each other.
    \item FreeKD-loss: our FreeKD  without using the reinforced knowledge judge. It determines the directions of knowledge distillation only relying on the cross entropy loss.
    \item FreeKD-all-neighbors: our FreeKD selecting the directions of node-level knowledge distillation via node-level actions, but using all neighborhood nodes as the local structure.
    \item FreeKD-all-structures: our FreeKD selecting the directions of node-level knowledge distillation, but without using structure-level actions for  structure-level knowledge distillation.
\end{itemize}

Table \ref{ablation_study} shows the  results. 
FreeKD-node is better than GCN, showing that mutually transferring node-level knowledge via reinforcement learning is useful for boosting the performance of GNNs.
FreeKD obtain better results than FreeKD-node. It illustrates distilling structure knowledge by our method is beneficial to GNNs.
FreeKD achieves better performance than FreeKD-w.o.-judge, illustrating dynamically determining the knowledge distillation direction is important.
In addition, FreeKD outperforms FreeKD-loss. This shows that directly using the cross entropy loss to decide the directions of knowledge distillation is sub-optimal. As stated before, this heuristic strategy only considers the performance of the node itself, but neglects the influence of the node on other nodes.
Additionally, FreeKD has superiority over FreeKD-all-neighbors, demonstrating that transferring part of neighborhood information selected by our method is more effective than transferring all neighborhood information for GNNs. 
Finally, FreeKD obtains better performance than FreeKD-all-structures, which indicates our reinforcement learning based method can transfer more reliable structure-level knowledge. 
In summary, these results demonstrate our proposed knowledge distillation framework with a hierarchical reinforcement learning strategy is effective.

\begin{table}
\caption{Ablation Study on the Chameleon and Cora dataset.}
\vspace{-0.1in}
 \setlength{\tabcolsep}{1.2mm}{
\begin{tabular}{@{}ccccccc@{}}
\toprule
& & &\multicolumn{2}{c}{Chameleon} &\multicolumn{2}{c}{Cora} \\ 
\hline
\multicolumn{1}{c|}{Method}                 & \multicolumn{2}{c|}{Network}     & \multicolumn{2}{c|}{F1 Score}     & \multicolumn{2}{c}{F1 Score}  \\
\multicolumn{1}{c|}{}                       & $\Phi$ & \multicolumn{1}{c|}{$\Psi$} & $\Phi$  & \multicolumn{1}{c|}{$\Psi$}    & $\Phi$  & \multicolumn{1}{c}{$\Psi$}   \\
\hline
\multicolumn{1}{c|}{GCN} & - & \multicolumn{1}{c|}{-} & 33.09 &\multicolumn{1}{c|}{-}  & 85.12 & - \\
\multicolumn{1}{c|}{FreeKD-node} & GCN & \multicolumn{1}{c|}{GCN} & 36.35 & \multicolumn{1}{c|}{36.42}  & 86.17 & 86.03 \\
\hline
\multicolumn{1}{c|}{FreeKD-w.o.-judge}     & GCN  & \multicolumn{1}{c|}{GCN}  & 35.33 & \multicolumn{1}{c|}{35.27}          & 85.83 & 85.76  \\
\multicolumn{1}{c|}{FreeKD-loss} & GCN  & \multicolumn{1}{c|}{GCN}  & 35.86 & \multicolumn{1}{c|}{35.79}           & 85.89 & 85.97 \\
\multicolumn{1}{c|}{FreeKD-all-neighbors} & GCN  & \multicolumn{1}{c|}{GCN}  & 36.85 & \multicolumn{1}{c|}{36.73}      & 86.21 & 86.26 \\
\multicolumn{1}{c|}{FreeKD-all-structures} & GCN  & \multicolumn{1}{c|}{GCN}  & 36.53 & \multicolumn{1}{c|}{36.62}     & 86.13 & 86.07  \\
\multicolumn{1}{c|}{FreeKD}                  & GCN  & \multicolumn{1}{c|}{GCN}  & 37.61 & \multicolumn{1}{c|}{37.70}    & 86.53 & 86.62 \\ \bottomrule
\end{tabular}
}
 \label{ablation_study}
  \vspace{-0.15in}
\end{table}
\subsection{Visualizations}
    We further intuitively show the effectiveness of the reinforced knowledge judge to dynamically decide the directions of knowledge distillation.
    We set GCN as $\Phi$ and GraphSAGE as $\Psi$, and train our FreeKD on the Cora dataset.
    Then, we poison  $\Phi$ by adding random Gaussian noise with a standard deviation $\sigma$ to its model parameters. Finally, we visualize the agent’s output, i.e., node-level policy probabilities $\pi_\mathbf{\theta}\left(\mathbf{s}_i^{[1]},0\right)$  and $\pi_\mathbf{\theta}\left(\mathbf{s}_i^{[1]},1\right)$ at node $i$ for $\Phi$ and $\Psi$, respectively. To better visualize, we show a subgraph composed of the first 30 nodes and their neighborhoods.

    Figure \ref{vis} shows the results using different standard deviations $\sigma$. 
    In Figure \ref{vis} (a), (b), and (c), the higher the probability output by the agent is, the redder the node is. And this means that the probability for the node in this network to serve as a distilled node to transfer knowledge to the corresponding node of the other network is higher.
	As shown in Figure \ref{vis} (a), when without adding noise, the degrees of the red color in $\Phi$ and $\Psi$ are comparable. As the noise is gradually increased in $\Phi$, the red color becomes more and more light in $\Phi$, but an opposite case happens in $\Psi$, as shown in Figure \ref{vis} (b) and (c). This is because the noise brings negative influence on the outputs of the network, leading to inaccurate soft labels and large losses. In such a case, our agent can output low probabilities for the network $\Phi$.  Thus, our agent can effectively determine the direction of knowledge distillation for each node.

% 	Figure \ref{vis}(a) visualize the result of add no Gaussian noise to net 1 GCN, where the degree of red for the whole subgraph are comparable. And for different nodes and local structures, GCN and GraphSAGE show different degree of red, because the agent could judge which GNN learns better and guide the bidirectional knowledge distillation between two GNNs at node-level and structure-level. Figure 3(b) and Figure 3(c) show the results of add Gaussian noise only to the parameters of net 1 GCN with standard deviation $\sigma=0.5$ and $\sigma=1.0$, respectively. We can observe that after adding the perturbations to poison net 1 GCN, net 2 GraphSAGE get redder and net 1 GCN become less red for the whole subgraph. The bigger the standard deviation $\sigma$ is, the redder net 2 GraphSAGE is. And when $\sigma=1.0$, net 1 GCN become almost totally yellow for the whole subgraph, since in this circumstance knowledge from net 1 GCN are almost harmful and not reliable. These visualizations indicates that the agent could commendably judge between beneficial knowledge and harmful knowledge, and measure how much one GNN learned relative better for node $i$ than another GNN, thus could effectively guide the bidirectional knowledge distillation between two GNNs.

\begin{table}
\caption{Sensitivity study of $\gamma$ on the Cora dataset.}
\vspace{-0.1in}
\setlength{\tabcolsep}{0.9mm}{%
\begin{tabular}{@{}c|cc|cccccc@{}}
\toprule
\multirow{2}{*}{Dataset} & \multicolumn{2}{c|}{Network} & \multirow{2}{*}{$\gamma$=0.0} & \multirow{2}{*}{$\gamma$=0.1} & \multirow{2}{*}{$\gamma$=0.3} & \multirow{2}{*}{$\gamma$=0.5} & \multirow{2}{*}{$\gamma$=0.7} & \multirow{2}{*}{$\gamma$=0.9} \\
                         & $\Phi$          & $\Psi$         &                        &                        &                        &                        &                        &                        \\ \midrule

\multirow{2}{*}{Cora}    & GCN           & GCN          & 86.32                  & 86.39                  & \textbf{86.57}         & 86.31                  & 86.12                  & 85.23                   \\
                         & GAT           & GAT          & 86.21                  & 86.45                  & 86.41                  & \textbf{86.57}         & 86.32                  & 86.22                  \\ \bottomrule
\end{tabular}%
}
 \label{gamma}
 \vspace{-0.1in}
\end{table}

\subsection{ Sensitivity and Convergence Analysis}
\begin{figure}
\centering
\subfigure[Sensitivity study of $\mu$ and $\rho$.]{
\begin{minipage}[t]{0.49\linewidth}
\centering
\includegraphics[width=1.56in]{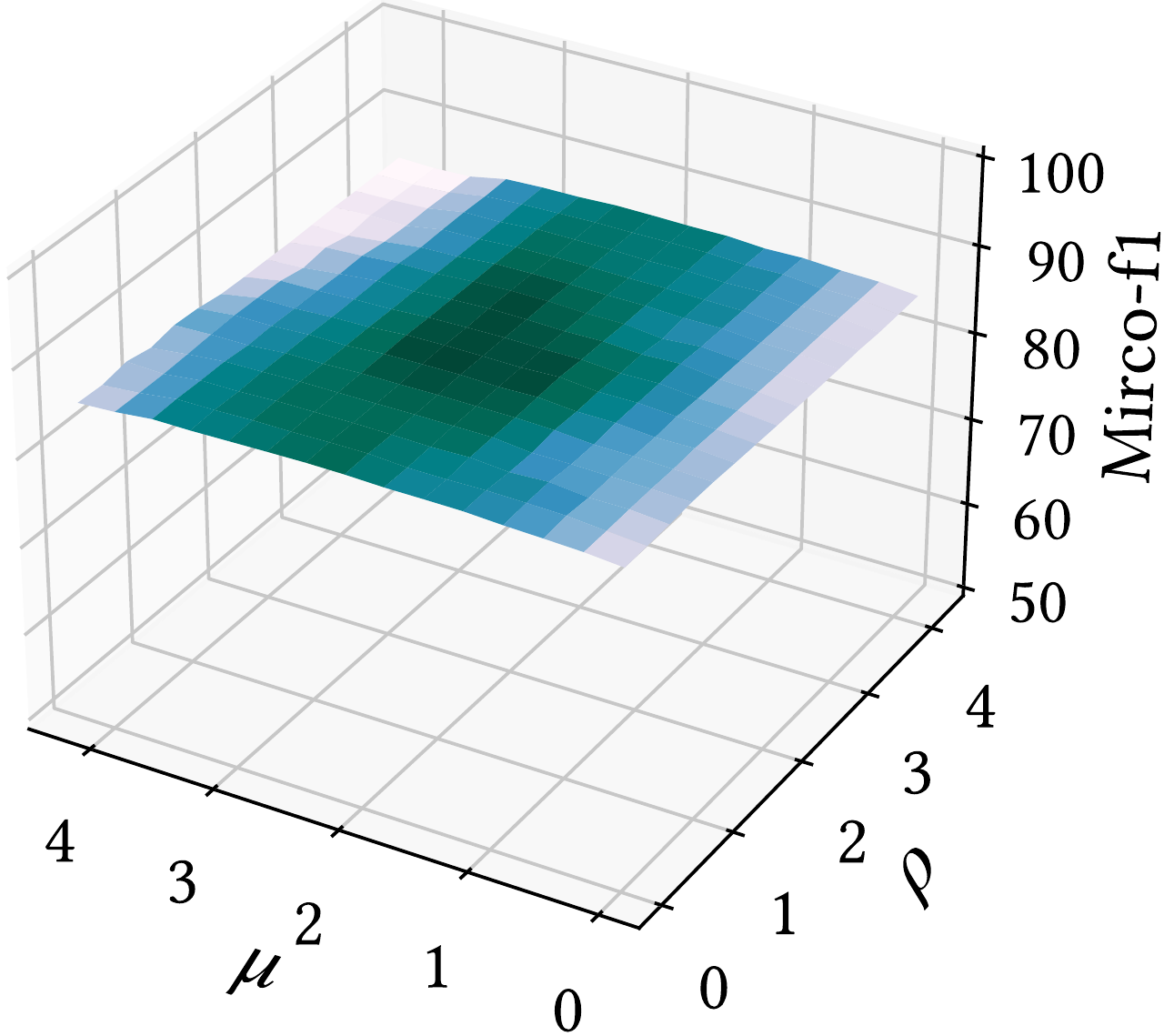}
\end{minipage}%
}%
\subfigure[Convergence curve.]{
\begin{minipage}[t]{0.49\linewidth}
\centering
\includegraphics[width=1.56in]{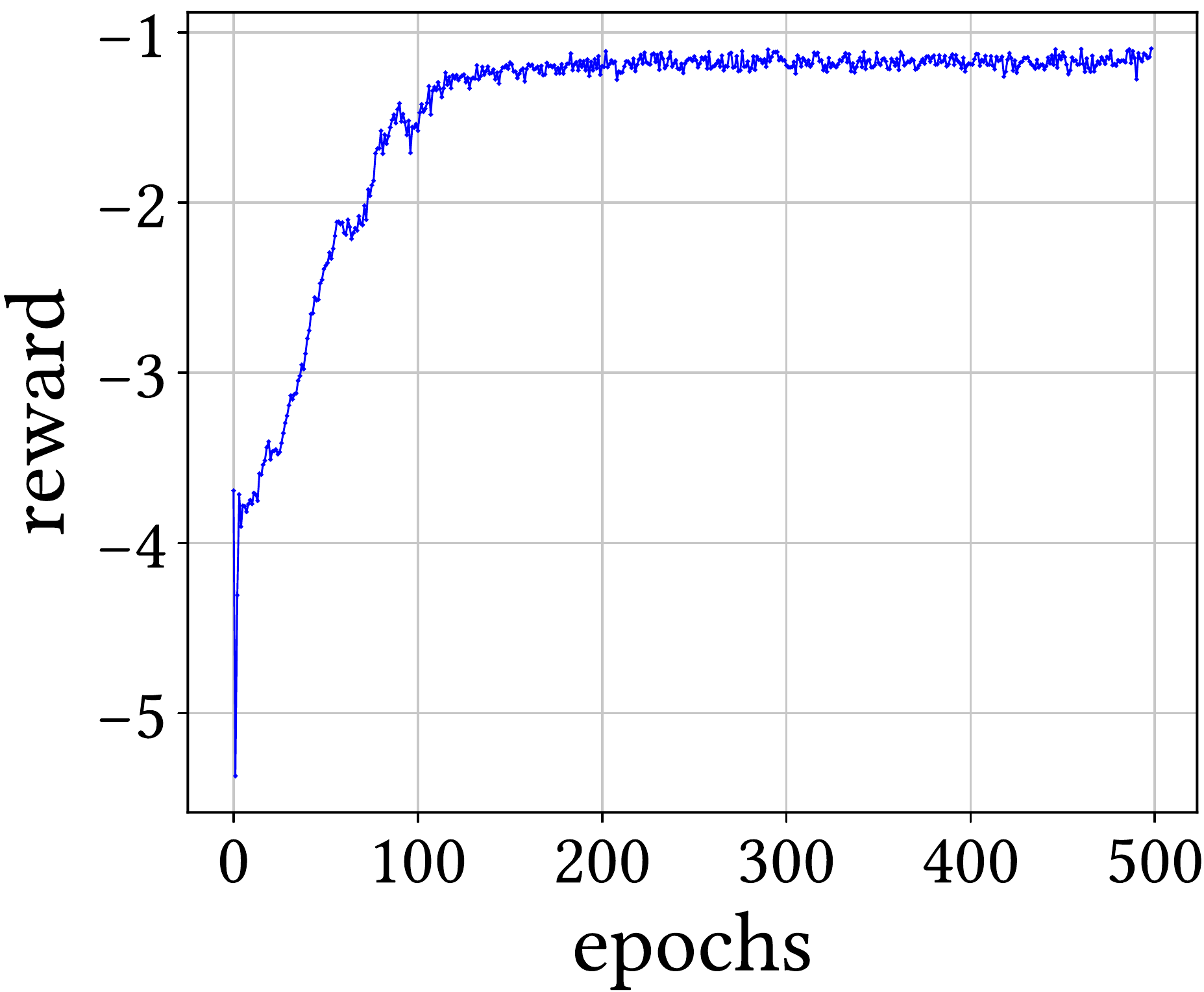}
\end{minipage}%
}%
\centering
\vspace{-0.2in}
\caption{Sensitivity study of $\mu$,$\rho$ and convergence analysis on the Cora dataset.}
\label{param}
\vspace{-0.17in}
\end{figure}

In our method, there are three main hyper-parameters, i.e., $\gamma$ in the reward function (\ref{reward}), $\mu$ and $\rho$ in the loss function (\ref{loss1}) and (\ref{loss2}). We study the sensitivity of our method to these hyper-parameters on the Cora dataset.
First, we investigate the impact of $\gamma$ in the agent’s reward function on the performance of our method. 
As shown in Table \ref{gamma}, with the values of $\gamma$ increasing, 
the performance of our method will fall after rising. In the meantime, our method is not sensitive to $\gamma$ in a relatively large range.
We also study the parameter sensitiveness of our method to $\mu$ and $\rho$. 
Figure \ref{param}(a) shows the results. Our method is still not sensitive to these two hyper-parameters in a relatively large range.
Additionally, we analyze the convergence of our method. Figure \ref{param}(b) shows the reward convergence curve. Our method is convergent after 100 epochs.
% Thus, it is easy to set the values of these hyper-parameters in real-world applications. 

% \begin{figure}[h]
%   \centering
%   \includegraphics[width=0.45\linewidth]{figure/param_cora3}
%   \vspace{-0.1in}
%   \caption{Sensitivity study of $\mu$ and $\rho$ on the Cora dataset. 
%   }
%   \label{param}
%   \vspace{-0.18in}
% \end{figure}

% Please add the following required packages to your document preamble:
% \usepackage{booktabs}
% \usepackage{multirow}
% \usepackage{graphicx}

% 	Recall that $\mu$ denotes the contribution factor of node-level distillation loss to the whole loss, while $\rho$ denotes the contribution factor of structure-level distillation loss to the whole loss. 
%	We visualize the mean F1-score of two GCN models training by BKDR with different hyper-parameters $\mu$ and $\rho$. 
% 	We choose the $\mu$ and $\rho$ ranging from 0 to 4.5 with a interval of 0.3. Figure 4 shows the results. The darker the grid is, the higher mean F1-score the grid has. 
% 	As shown in Figure \ref{param}, we observe that increasing $\mu$ and $\rho$ in appropriate range improve the performance of GNNs, indicating that the agent-guided knowledge transfer at node-level and at structure-level both play significant roles in BKDR. 
%Tthe results. Our method is still not sensitive to these two hyper-parameters in a relatively large range. Thus, it is easy to set the values of these hyper-parameters in real-world applications. 
% 	However, too large or too small values of $\mu$ and $\rho$ should be avoided in practice, which could result in bad performance. Because too small values could make the bidirectional distillation contribute too little during training, while too large values could lead to the ineffectiveness of cross entropy loss with ground-truth label. 

\section{Conclusion}

In this paper, we proposed a free-direction  knowledge distillation framework to enable two shallower GNNs to learn from each other, without requiring  a deeper well-optimized teacher GNN. Meanwhile, we devised a hierarchical reinforcement learning  mechanism to manage the directions of knowledge transfer, so as to distill knowledge from both node-level and structure-level aspects. Extensive experiments demonstrated the effectiveness of our method.

%%
%% The acknowledgments section is defined using the "acks" environment
%% (and NOT an unnumbered section). This ensures the proper
%% identification of the section in the article metadata, and the
%% consistent spelling of the heading.
\begin{acks}
This  work  was  supported  by  the National Natural Science Foundation of China (NSFC) under Grants 62122013, U2001211.
\end{acks}

%%
%% The next two lines define the bibliography style to be used, and
%% the bibliography file.
\bibliographystyle{ACM-Reference-Format}
\bibliography{myreference}

%%
%% If your work has an appendix, this is the place to put it.
\clearpage
\appendix

\section{APPENDIX}
To support the reproducibility of our work, we introduce more details about training and experiments in the appendix.
\subsection{Training procedure}
\label{procedure}
Algorithm 1 is the pseudo-code of the our FreeKD training procedure. 
The GNNs and the agent closely interact with each other when training. For each batch, we first calculate the cross entropy loss for training GNNs. The node-level states for a batch of nodes are then calculated and feed into the node-level policy network. After that, we sample node-level actions from the policy probabilities produced by the agent to decide the directions of knowledge distillation between two GNNs. Then, the agent receives structure-level states from environment and produces structure-level actions to decide which of the local structures to be propagated.
The two-level states and actions are stored in the history buffer. Next, we train the two GNNs with the overall loss. After that, for the stored states and actions, we calculate the delayed rewards according to the performance of GNNs and update the policy network with gradient ascent. The GNNs and the agent are learned together and mutually improved.
\begin{algorithm}[h]
  \caption{Free-direction Knowledge Distillation Framework via Reinforcement Learning for GNNs}  
  \begin{algorithmic}[1]  
    \Require 
    graph $\mathbf{G=(V,E,X)}$, label set $\mathbf{\mathcal{Y}}$, epoch number $L$;
    \Ensure
    the predicted classes of nodes in the GNN models $\Phi$ and $\Psi$, the trained parameters of  $\Phi$ and  $\Psi$;
    \State Initialize  $\Phi$ and  $\Psi$;
    \State Initialize the policy networks in reinforcement learning;
    \For{each epoch $k$ in $1$ to $L$}  
        \For{each batch in epoch $k$}  
		    \State calculate the cross entropy losses $L_{CE}^\Phi$, $L_{CE}^\Psi$;
		    \State calculate node-level states for a batch of nodes by (\ref{n-state});
		    \State  sample node-level actions by $a_i^{[1]}\sim\pi_\mathbf{\theta}(\mathbf{s}_i^{[1]},a_i^{[1]})$;
		    \State derive structure-level states by (\ref{s-state});
		    \State  sample structure-level actions by $a_i^{[2]}\sim\pi_\mathbf{\delta}(\mathbf{s}_i^{[2]},a_i^{[2]})$;
		    \State store states and actions to history buffer $H$;
		    \State  calculate $L_{node}^\Phi$ and $L_{node}^\Psi$  by (\ref{ce1}) and (\ref{ce2});
		    \State calculate $L_{struct}^\Phi$ and $L_{struct}^\Psi$  by (\ref{struct1}) and (\ref{struct2});
		    \State  calculate the overall losses $L^\Phi$, $L^\Psi$ by (\ref{loss1}) and (\ref{loss2});
		    \State update the parameters of  $\Phi$ by minimizing $L^\Phi$;
		    \State update the parameters of $\Psi$ by minimizing $L^\Psi$;
		    \For{each state and action in buffer $H$}
            	\State calculate the delayed rewards by (\ref{reward});
        		\State update parameters of the policy networks by (\ref{update});
            \EndFor
        \EndFor  
    \EndFor     
  \end{algorithmic}  
\end{algorithm}  

\subsection{Datasets Description and Statistics}
\label{datasets}
In our experiments, we use five widely used public datasets of different domains to evaluate our method. Table \ref{dataset} summarizes the statistics of the five datasets. In the following we introduce more information about these five datasets.

\textbf{Cora \cite{sen2008collective} and Citeseer \cite{sen2008collective}} are two citation networks. Cora is composed of papers in the machine learning domain, and Citeseer is about computer science publications. In these two datasets, nodes stand for documents while edges represent the citation relationship between documents. The node feature is the bag-of-words representation of document and the node label is the corresponding research domain.

\textbf{Chameleon \cite{rozemberczki2021multi}  and Texas \cite{pei2020geom}} are two web networks. In these two datasets, the nodes represent web pages and the edges are the hyperlinks between web pages. The node feature is the bag-of-words vector that represents the corresponding web page and the node label is its corresponding category.

\textbf{PPI \cite{hamilton2017inductive}} describes the protein-protein interactions in human tissues. PPI is a widely used inductive learning dataset,  containing 24 protein-protein interactions graphs. Every graph is about a specific human tissue. PPI takes positional gene sets, motif gene sets and immunological signatures as the node feature, and takes the corresponding gene ontology set as the node label. 

\begin{table}[h]
\vspace{-0.1in}
\caption{Dataset statistics.}
\vspace{-0.1in}
 \setlength{\tabcolsep}{1.3mm}{
\begin{tabular}{@{}cccccc@{}}
\toprule
Dataset   & \# Graphs & \# Nodes & \# Edges & \# Features & \# Classes \\ \midrule
Cora      & 1         & 2708     & 5429     & 1433        & 7          \\
Citeseer  & 1         & 3327     & 4732     & 3703        & 6          \\
Chameleon & 1         & 2277     & 36101    & 2325        & 4          \\
Texas     & 1         & 183      & 309      & 1703        & 5          \\
PPI       & 24        & 56944    & 818716   & 50          & 121        \\ \bottomrule
\end{tabular}
}
\vspace{-0.2in}
 \label{dataset}
\end{table}

\subsection{Implementation Details}
\label{details}
 All the results are averaged above 10 times and we run our experiments on GeForce RTX 2080 Ti GPU.
 We use the Adam optimizer \cite{kingma2014adam} for training and adopt early stopping with a patience on validation sets of $150$ epochs. The initial learning rate is $0.05$ for GAT and  $0.01$ for GCN, GraphSAGE, and is decreased by multiplying $0.1$ every $100$ epochs. For the reinforced knowledge judge module, we set a fixed learning rate of $0.01$. We set the dropout rate to $0.5$ and the $l_2$ norm regularization weight decay to $0.0005$. The parameters of all GNN models are randomly initialized.  
The hyper-parameters  $\mu$ and $\rho$ in our method are searched from $\{0.5, 1.0,1.5,2.0\}$, and $\gamma$ is searched from $\{0.1,0.3,0.5\}$. The node-level policy network and structure-level policy network are both 3-layer MLP with tanh activation function and the size of hidden layer is set to $\{64,32\}$.
For the transductive setting, the number of layers in GNNs is set to $2$ and
the hidden size is set to $64$. For the inductive setting, the number
of layers in GNNs is set to $3$ and the hidden size is set to $256$. For GAT, the attention dropout probability is set to $0.5$ and the number of attention heads is set to $8$. For GraphSAGE, we use the mean aggregator to sample neighbors. 
% For GIN, the training epsilon is fixed to $0$ for better generalization as the original paper suggests. For APPNP, the propagation iteration steps is set to $8$ and the teleport probability is set to $0.2$. For SGC, the propagation iteration steps is set to $4$.

In the experiments of comparison with other knowledge distillation methods, the student model, i.e., GAT, is set to 2-layer with 64 hidden size in the transductive setting and 3-layer with 256 hidden size in the inductive setting.
For the teacher model GCNII, the number of layers is set to $32$ and the hidden size is set to $128$ in the transductive setting; in the inductive setting, the number of layers is set to $9$ and the hidden size is set to $2048$.
For all the compared knowledge distillation methods, we use the parameters as their original papers suggest and report their best results.

\end{document}